%% file: naaclhlt2019.tex
\documentclass[11pt,a4paper]{article}

\usepackage{times}
\usepackage{latexsym}
\usepackage{amsmath}
\usepackage{soul}
\usepackage{color}
\usepackage{array}
\usepackage[utf8]{inputenc} %
\usepackage[T1]{fontenc}    %
\usepackage{url}            %
\usepackage{booktabs}       %
\usepackage{amsfonts}       %
\usepackage{nicefrac}       %
\usepackage{microtype}      %
\usepackage{multicol}
\usepackage{multirow}
\usepackage{cancel}
\usepackage{caption}
\usepackage{cabin}
\usepackage{subcaption}
\usepackage{xspace}
\usepackage{graphicx}
\usepackage{siunitx}
\usepackage{amsfonts}
\usepackage{makecell}
\usepackage{comment}
\usepackage[inline]{enumitem}
\usepackage{hyperref}
\usepackage{listings}
\usepackage[svgnames]{xcolor}
\usepackage{bold-extra}
\usepackage{etoolbox}
\hypersetup{
    colorlinks=true,
    linkcolor=blue,
    filecolor=magenta,      
    urlcolor=blue,
}

\newtoggle{arxiv}
\toggletrue{arxiv}

\iftoggle{arxiv}
{
\definecolor{linkblue}{RGB}{5,0,127}
\usepackage[sort]{natbib}
\hypersetup{colorlinks=true,allcolors=linkblue}
\setlength{\textwidth}{6.5in}
\setlength{\textheight}{9in}
\setlength{\oddsidemargin}{0in}
\setlength{\evensidemargin}{0in}
\setlength{\topmargin}{-0.5in}
\newlength{\defbaselineskip}
\setlength{\defbaselineskip}{\baselineskip}
\setlength{\marginparwidth}{0.8in}
}

\definecolor{codegreen}{rgb}{0,0.6,0}
\definecolor{codegray}{rgb}{0.5,0.5,0.5}
\definecolor{codepurple}{rgb}{0.58,0,0.82}
\definecolor{backcolour}{rgb}{0.95,0.95,0.92}

\lstdefinestyle{inlinestyle}{
    backgroundcolor=\color{backcolour},   
    commentstyle=\color{codegreen},
    keywordstyle=\color{magenta},
    numberstyle=\tiny\color{codegray},
    stringstyle=\color{codepurple},
    basicstyle=\ttfamily\scriptsize\noindent,
    breakatwhitespace=false,         
    breaklines=true,                 
    captionpos=b,
    xleftmargin=-0.4cm,
    framexleftmargin=0cm,
    framexrightmargin=0cm,
    belowskip=0.2cm,
    aboveskip=0.3cm,
    keepspaces=true,                 
    numbers=left,                    
    numbersep=5pt,                  
    showspaces=false,                
    showstringspaces=false,
    showtabs=false,                  
    tabsize=2,
    frame=single,
    language=Python,
    numbers=none,
}

\lstdefinestyle{mystyle}{
    backgroundcolor=\color{backcolour},   
    commentstyle=\color{codegreen},
    keywordstyle=\color{magenta},
    numberstyle=\tiny\color{codegray},
    stringstyle=\color{codepurple},
    basicstyle=\ttfamily\scriptsize\noindent,
    breakatwhitespace=false,         
    breaklines=true,                 
    captionpos=b,
    xleftmargin=0.1cm,
    framexleftmargin=0cm,
    framexrightmargin=0cm,
    belowskip=0cm,
    aboveskip=0cm,
    keepspaces=true,                 
    numbers=left,                    
    numbersep=5pt,                  
    showspaces=false,                
    showstringspaces=false,
    showtabs=false,                  
    tabsize=2,
    frame=single,
    language=Python,
    numbers=none,
}

\definecolor{ctemplate}{rgb}{0.23, 0.30, 0.45}
\definecolor{cword}{rgb}{0, 0, 0.7}
\newcommand{\template}[1]{\texttt{\textcolor{ctemplate}{``#1''\xspace}}}
\newcommand{\ttag}[1]{\texttt{\textcolor{cword}{\{\MakeUppercase{#1}\}\xspace}}}
\usepackage{tikz}
\newcommand{\cbox}[1]{
    \begin{tikzpicture}
    \path[draw=#1,fill=#1] (0,0) rectangle (1.5mm,1.5mm);
    \end{tikzpicture}
}
\definecolor{attack}{HTML}{FFC253}
\definecolor{aug}{HTML}{FF854D}
\definecolor{sp}{HTML}{86A8FF}
\definecolor{eval}{HTML}{E75C9C}

\newcommand*\samethanks[1][\value{footnote}]{\footnotemark[#1]}
\DeclareMathOperator*{\argmax}{arg\,max} %

\title{\raisebox{-0.2cm}{\largerglogo}\, \RG{}: Unifying the NLP Evaluation Landscape}

\iftoggle{arxiv}{
\usepackage{authblk}

\author[1]{Karan Goel\thanks{Equal contribution. KG, NR, and JV made significant contributions.}}
\author[2]{Nazneen Rajani\samethanks}
\author[2]{Jesse Vig}
\author[2]{\\ Samson Tan}
\author[2]{Jason Wu}
\author[2]{Stephan Zheng} 
\author[2]{Caiming Xiong}
\author[3]{Mohit Bansal}
\author[1]{Christopher R\'{e}}
\affil[1]{Stanford University}
\affil[2]{Salesforce Research}
\affil[3]{UNC-Chapel Hill}
\affil[1]{\texttt{\{kgoel,chrismre\}@cs.stanford.edu}}
\affil[2]{\texttt{\{nazneen.rajani,jvig\}@salesforce.com}}
 \affil[3]{\texttt{\{mbansal\}@cs.unc.edu}}
\date{}
}

\newcommand\RG{Robustness Gym}
\newcommand{\RGabbrv}{{\rgfont RG}}

\makeatletter
\newcommand*{\rom}[1]{\expandafter\@slowromancap\romannumeral #1@}

\newcommand{\para}[1]{\vspace{0.5em}\noindent{\bf #1}}
\newcommand{\tightpara}[1]{\noindent{\bf #1}}

\newcommand{\bootleg}{{\sc Bootleg}}
\newcommand{\microsoft}{{\sc Microsoft}}
\newcommand{\google}{{\sc Google}}
\newcommand{\amazon}{{\sc Amazon}}
\newcommand{\wat}{{\sc Wat}}
\newcommand{\rel}{{\sc Rel}}
\newcommand{\pop}{{\sc Pop}}
\newcommand{\wiki}{{\sc Wikipedia}}
\newcommand{\aida}{{\sc Aida}}

\usepackage{bbold}
\newcommand*{\rgfont}{\bbfamily\selectfont}%

\newcommand{\rglogo}{%
  \begingroup
  \raisebox{-0.07cm}{
  \includegraphics[height=0.4cm]{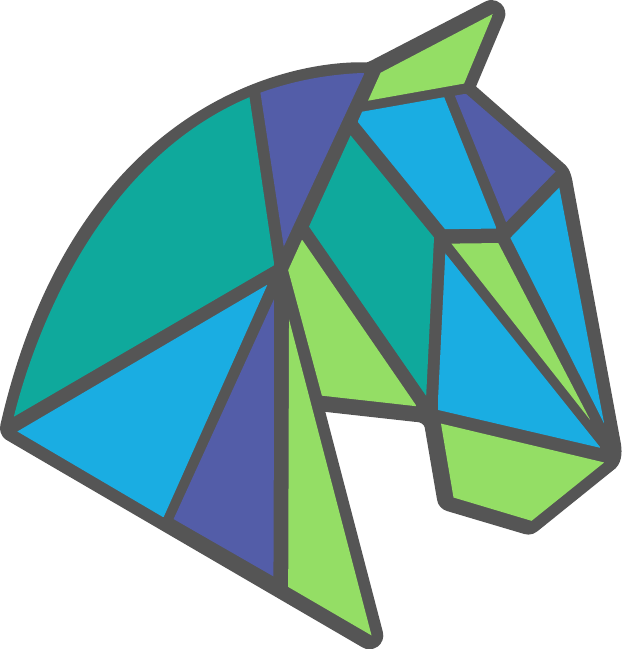}%
  }
  \endgroup
}

\newcommand{\largerglogo}{%
  \begingroup
  \includegraphics[height=0.9cm]{figs/RG_LOGO.pdf}%
  \endgroup
}

\newcommand{\stylizeconcepts}{\cabin}
\newcommand{\slicebuilder}{{\stylizeconcepts SliceBuilder}}
\newcommand{\slicebuilders}{{\stylizeconcepts SliceBuilder}s}
\newcommand{\cachedop}{\mbox{{\stylizeconcepts CachedOperation}}}
\newcommand{\cachedops}{{\stylizeconcepts CachedOperation}s}
\newcommand{\testbench}{{\stylizeconcepts TestBench}}

\newcommand{\report}{{\stylizeconcepts Report}}

\makeatother
\begin{document}

\maketitle

\begin{abstract}

Despite impressive performance on standard benchmarks, deep neural networks are often brittle when deployed in real-world systems. 
Consequently, recent research has focused on testing the robustness of such models, resulting in a diverse set of evaluation methodologies ranging from adversarial attacks to rule-based data transformations. 
In this work, we identify challenges with evaluating NLP systems and propose a solution in the form of \mbox{\rglogo\ {\RG{}} (\RGabbrv)},\footnote{\,\url{https://robustnessgym.com/}} a simple and extensible evaluation toolkit that unifies $4$ standard evaluation paradigms: subpopulations, transformations, evaluation sets, and adversarial attacks. 
By providing a common platform for evaluation, \RG{} enables practitioners to compare results from all $4$ evaluation paradigms with just a few clicks, and to easily develop and share novel evaluation methods using a built-in set of abstractions.
To validate \RG{}'s utility to practitioners, we conducted a real-world case study with a sentiment-modeling team, revealing performance degradations of $18\%+$. 
To verify that \RG{} can aid novel research analyses, we perform the first study of state-of-the-art commercial and academic named entity linking (NEL) systems, as well as a fine-grained analysis of state-of-the-art summarization models.
For NEL, commercial systems struggle to link rare entities and lag their academic counterparts by $10\%+$, while state-of-the-art summarization models struggle on examples that require abstraction and distillation, degrading by $9\%+$.

\end{abstract}

\input{source/intro}

\input{source/motivation_challenges}

\input{source/3cs}

\input{source/contemplate}
\begin{figure*}
    \centering
    \includegraphics[width=\textwidth]{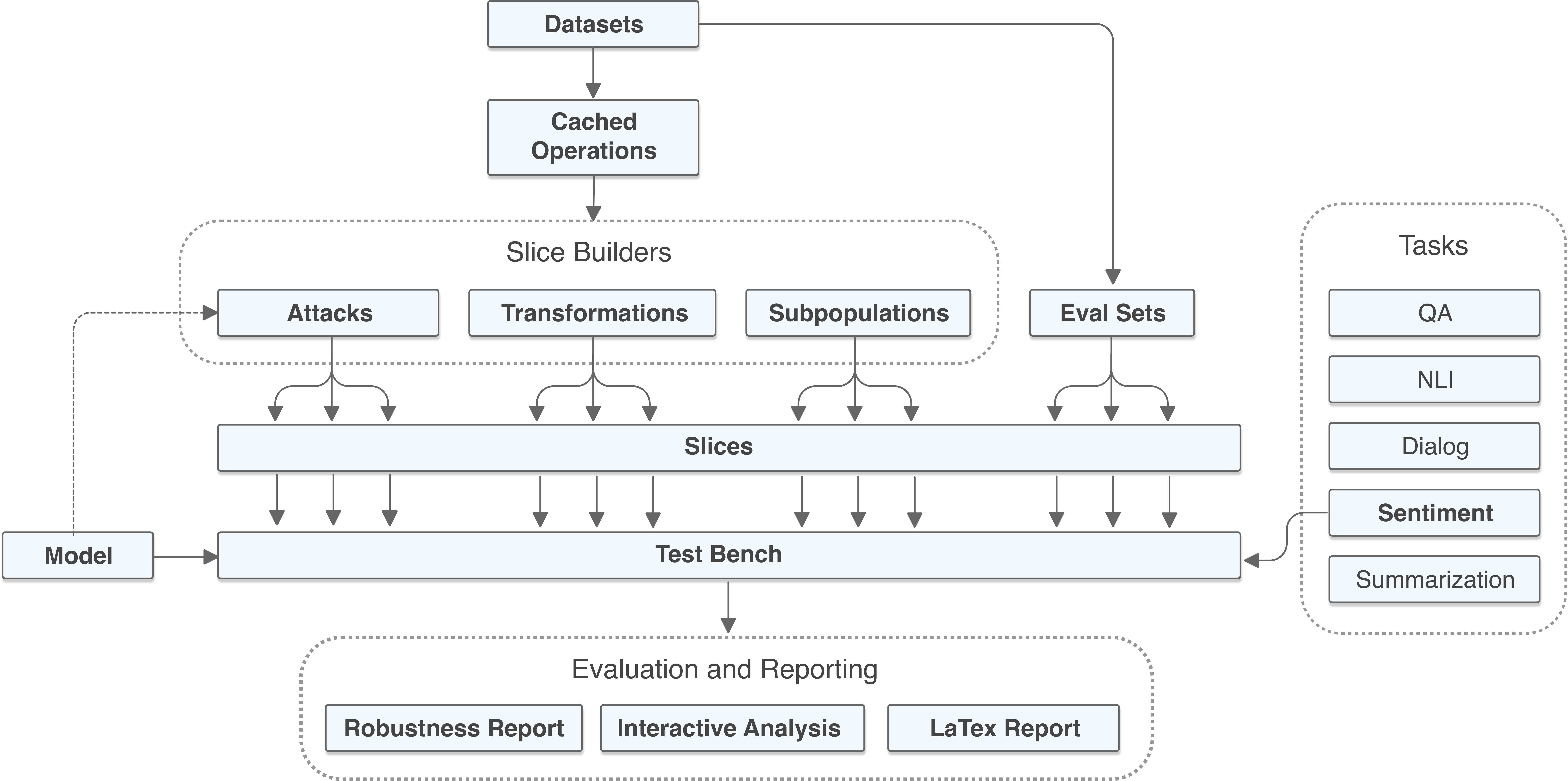}
    \caption{\RG~system design and workflow. 
    }
    \label{fig:rg-system-design}
\end{figure*}
\input{source/create}

\input{source/consolidate}

\begin{table*}[!t]
    \centering
    \scriptsize
    \begin{tabular}{llll}
\toprule
 &Type &Instantiation&Examples\\
\midrule
\multirow{7}{*}{\rotatebox[origin=c]{90}{Rule-based}}
&\multirow{3}{*}{Filters}
&HasPhrase&\cbox{sp} Subpopulation that contains negation. \\
&&HasLength& \cbox{sp} Subpopulation that is in the \ttag{x} percentile for length. \\
&&Position& \cbox{sp} Subpopulation that contains \ttag{token} in position \ttag{n}. \\
\cmidrule{3-4}
&\multirow{3}{*}{Logic}
&IFTTT recipes& \cbox{aug} If example ends in \ttag{ing} then transform with backtranslation. \\
&&Symmetry& \cbox{eval} Switch the first and last sentences of a source document to create a new eval set. \\
&&Consistency& \cbox{attack} Adding ``aaaabbbb" at the end of every example as a form of attack. \\
\cmidrule{3-4}
&\multirow{1}{*}{Template}
&Checklist&\cbox{eval} Generate new eval set using examples of the form \template{I \ttag{NEGATION} \ttag{POS\_VERB}.} \\
\midrule
\multirow{15}{*}{\rotatebox[origin=c]{90}{Machine}}
&\multirow{2}{*}{Classifier}
&HasScore& \cbox{sp} Subpopulation with perplexity \ttag{>X} based on a LM.  \\
&&HasTopic& \cbox{sp} Subpopulation belonging to a certain topic. \\
\cmidrule{3-4}
&\multirow{2}{*}{Tagger*}
&POS& \cbox{sp} Subpopulation that contains \ttag{POS\_NOUN} in position \ttag{n}. \\
&&NER& \cbox{sp} Subpopulation that contains entity names with non-English origin.  \\
&&SRL& \cbox{sp} Subpopulation where there is no \ttag{agent}. \\
&&Coref& \cbox{sp} Subpopulation that contains the pronouns for a particular gender. \\
\cmidrule{3-4}
&\multirow{2}{*}{Parser*}
&Constituency&\cbox{aug} Transform with all complete subtrees of \ttag{POS\_VP} in the input. \\
&&Dependency& \cbox{sp} Subpopulation that has at least 2 \ttag{POS\_NP} dependent on \ttag{POS\_VP}.  \\
\cmidrule{3-4}
&\multirow{2}{*}{Generative}
& Backtranslation&\cbox{aug} Using a seq2seq model for transformation using backtranslation. \\
&&Few-shot& \cbox{eval} Using GPT3 like models for creating synthetic eval sets.\\
\cmidrule{3-4}
&\multirow{2}{*}{Perturbation}
&Paraphrasing&\cbox{aug} Synonym substitution using EDA. \\
&&TextAttack&\cbox{attack} Perturbing input using TextAttack recipes. \\
\midrule
\multirow{6}{*}{\rotatebox[origin=c]{90}{\parbox{2.2cm}{Human or \\ Human-in-the-loop}}}
&\multirow{1}{*}{Filtering}&Figurative text&\cbox{sp} Using humans to identify subpopulation that contain sarcasm. \\
\cmidrule{3-4}
&\multirow{2}{*}{Curation}
&Evaluation sets&\cbox{eval} Building datasets like ANLI, Contrast sets, HANS, etc. \\
&&Data validation&\cbox{aug} Using human-in-the-loop for label verification. \\
\cmidrule{3-4}
&\multirow{2}{*}{Adversarial}
&Invariant&\cbox{attack} Perturbing text in a way that the expected output does not change. \\
&&Directional&\cbox{attack} Perturbing text in a way that the expected output changes. \\
\cmidrule{3-4}
&\multirow{1}{*}{Transformation}
&Counterfactual&\cbox{aug} Transforming to counterfactuals for a desired target concept. \\
\bottomrule
    \end{tabular}
    \caption{Sample of slice builders and corresponding data slices along with example use cases that can either be used out-of-the-box or extended from \RG{}. \protect\cbox{sp} $\rightarrow$ subpopulations, \protect\cbox{eval} $\rightarrow$ evaluation sets, \protect\cbox{aug} $\rightarrow$ transformations and \protect\cbox{attack} $\rightarrow$ adversarial attacks. $*$ marked are \cachedops\ and the rest are \slicebuilders.}
    \label{tab:rg-slicebuilders}
\end{table*}

\input{source/personas}

\input{source/novice}
\input{source/intermediate}
\input{source/expert}

\input{source/experiments/commercial}

\input{source/experiments/experiments}

\input{source/experiments/ned}

\input{source/experiments/summarization}

\input{source/relatedwork}
\input{source/conclusion}

\section*{Acknowledgements}
This work was part of a collaboration between Stanford, UNC, and Salesforce Research and was supported by Salesforce AI Research grants to MB and CR. KG and NR conceived the idea of Robustness Gym. KG, NR, and JV made significant overall contributions to the toolkit. ST and JW ran initial experiments on some NLP tasks. SZ and CX provided useful feedback. MB and CR provided detailed guidance on the NLP/robustness and MLSys areas, respectively. 
We are thankful to Han Guo, Laurel Orr, Jared Dunnmon, Chris Potts, Marco Tulio Ribeiro, Shreya Rajpal for helpful discussions and feedback.
\\
\noindent
CR also gratefully acknowledges the support of NIH under No. U54EB020405 (Mobilize), NSF under Nos. CCF1763315 (Beyond Sparsity), CCF1563078 (Volume to Velocity), and 1937301 (RTML); ONR under No. N000141712266 (Unifying Weak Supervision); the Moore Foundation, NXP, Xilinx, LETI-CEA, Intel, IBM, Microsoft, NEC, Toshiba, TSMC, ARM, Hitachi, BASF, Accenture, Ericsson, Qualcomm, Analog Devices, the Okawa Foundation, American Family Insurance, Google Cloud, Swiss Re, Total, the HAI-AWS Cloud Credits for Research program, and members of the Stanford DAWN project: Facebook, Google, and VMWare. The U.S. Government is authorized to reproduce and distribute reprints for Governmental purposes notwithstanding any copyright notation thereon. Any opinions, findings, and conclusions or recommendations expressed in this material are those of the authors and do not necessarily reflect the views, policies, or endorsements, either expressed or implied, of NIH, ONR, or the U.S. Government.

\iftoggle{arxiv}{\setcitestyle{numbers}}{}

\bibliography{naaclhlt2019}
\bibliographystyle{plainnat}
\input{source/appendix}

\end{document}

%% file: source/intro.tex
\section{Introduction}

Advances in natural language processing (NLP) have led to models that achieve high accuracy when train and test data are independent and identically distributed (i.i.d.).
However, analyses suggest that these models are not robust to data corruptions~\citep{Belinkov2018SyntheticAN}, distribution shifts~\citep{hendrycks2020many, Miller2020TheEO}, or harmful data manipulations~\citep{Jia2017AdversarialEF}, and they may rely on spurious patterns~\citep{mccoy2019right} for prediction.
In practice, these vulnerabilities limit successful generalization to unseen data and hinder deployment of trustworthy systems. 
A consequence of this is the proliferation of public-use systems that were later revealed to be systematically biased~\citep{Chloe2018Microsoft, Isobel2018Amazon, Nicolas2020Google, Karen2019Facebook, Will2019Apple, Isobel2020Twitter}, such as
recruiting tools biased against women.%

\begin{figure}[t!]
    \centering
    \includegraphics[width=\linewidth]{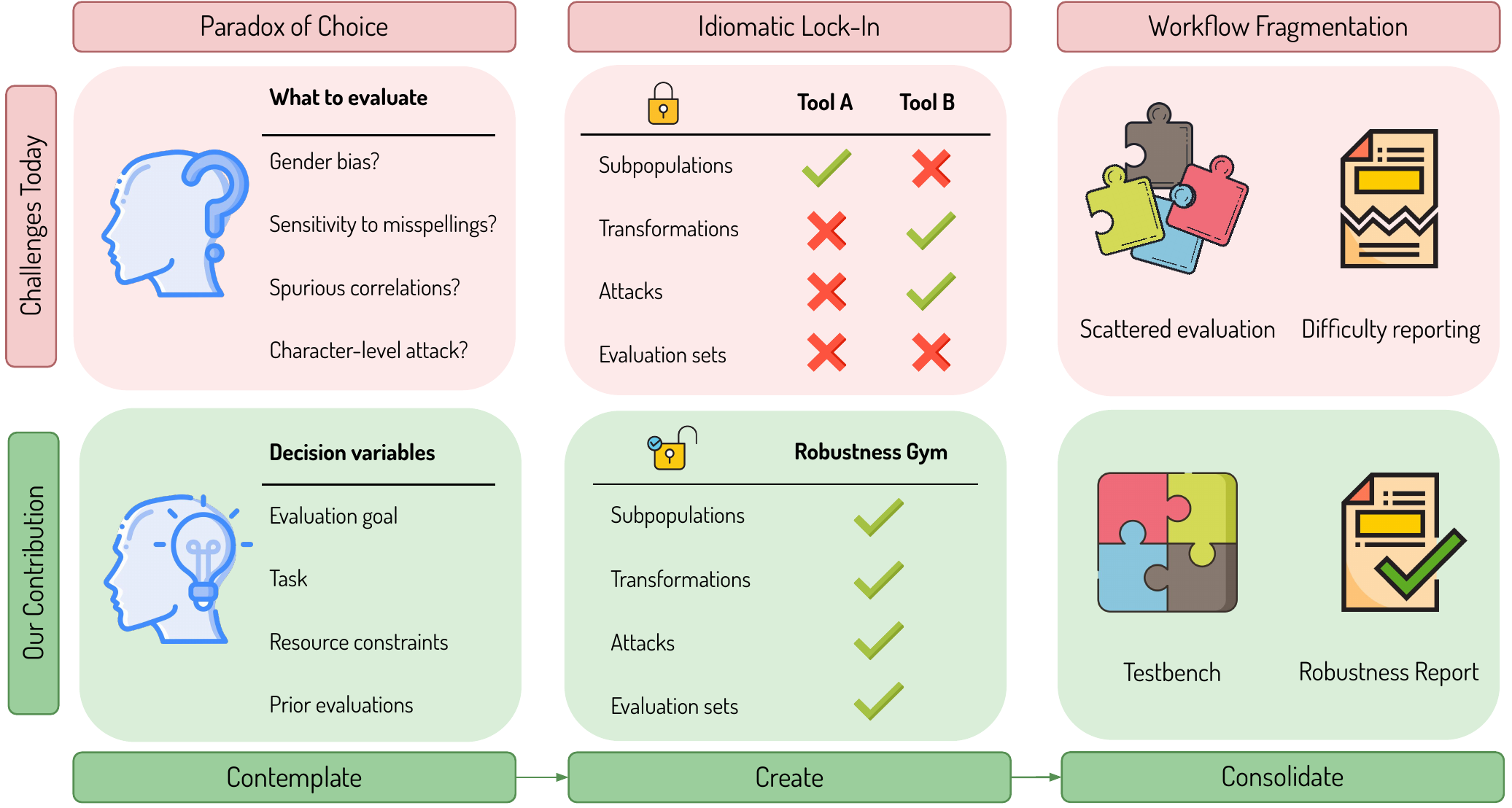}
    \caption{(top) challenges faced by practitioners evaluating their models, and (bottom) the Contemplate $\rightarrow$ Create $\rightarrow$ Consolidate evaluation loop uses \RG{} to address these challenges.}
    \label{fig:splash}
\end{figure}

\tightpara{}While researchers and practitioners are aware of these problems, it remains common practice to report performance solely on i.i.d. test data.
Ideally, evaluation would continually test a model's capabilities on examples that it is likely to see when deployed, rather than produce a static artifact of the model's performance. 
This process can be complex for practitioners, since there is no systematic method to prioritize what to evaluate, which evaluation methods to apply, and how to leverage or share the findings of previous evaluations. 
In summary, current evaluation practices face three challenges (Figure~\ref{fig:splash}, top): %

\begin{enumerate}[leftmargin=*]
    \item \textbf{Paradox of choice (Section~\ref{challenge:paradox_of_choice})}. 
    Even without writing a single line of code, practitioners are often confused about %
    what evaluations to run.
    This stems from a lack of prior research on how to evaluate, especially guidance that is sensitive to the practitioner's task, evaluation needs, and resource constraints.
    This confusion becomes a key challenge when treating evaluation as a continual process, since prior evaluation attempts and findings should influence a practitioner's future evaluation decisions.%

    \item \textbf{Idiomatic lock-in (Section~\ref{challenge:idiomatic_lock_in}).} 
    When determining the evaluation they want to run, practitioners must also choose an appropriate tool. %
    We identify $4$ distinct evaluation idioms supported by existing tools and research---subpopulations, transformations, adversarial attacks and evaluation sets. 
    Each tool uses bespoke abstractions to serve a subset of these idioms (e.g. adversarial attacks on words), requiring users to glue together multiple tools to perform a broad evaluation that mixes idioms.

    \item \textbf{Workflow fragmentation (Section~\ref{challenge:workflow_fragmentation}).} 
    As practitioners evaluate, they need to save progress, report findings and collaborate to understand model behavior. 
    Existing solutions to save progress are tool- and idiom-specific, lack versioning and provide limited support for sharing. 
    Existing reporting templates~\citep{mitchell2019model} are free-form, and have not successfully incentivized users to report findings e.g. we find only $6$\% of Huggingface~\citep{wolf-etal-2020-transformers} models have any evaluation information reported. %
    
\end{enumerate}

\tightpara{}In response to these challenges, we introduce \RG{} (\RGabbrv), a simple, extensible, and unified toolkit for evaluating robustness and sharing findings. We embed \RGabbrv{} in a new paradigm for continually evaluating models: the Contemplate $\rightarrow$ Create $\rightarrow$ Consolidate evaluation loop (Figure~\ref{fig:splash}, bottom). In this loop, we envision that researchers and practitioners will:
\begin{enumerate}[leftmargin=*]
    \item \textbf{Contemplate (Section~\ref{subsection:contemplate})} what evaluation to run next. %
    We provide guidance to practitioners on how key variables---their task, evaluation needs and resource constraints---can help prioritize which evaluation to run next. 
    We describe the influence of the task via the task schema and known prior evaluations, needs such as testing generalization, bias, or security, and constraints such as expertise, access to compute, and human resources. %
    We describe how these decisions could evolve as more evaluations are conducted. %
    
    \item \textbf{Create (Section~\ref{subsection:create})} \emph{slices} of data in \RGabbrv, where each slice defines a collection of examples for evaluation, built using one or a combination of evaluation idioms. %
    \RGabbrv{} supports users in a simple two-stage workflow, separating the storage of side information about examples (\cachedop), away from the nuts and bolts of programmatically building slices across all $4$ idioms using this information (\slicebuilder). %
    This workflow allows users to quickly implement new ideas, minimize boilerplate code, and seamlessly integrate existing tools.

    \item \textbf{Consolidate (Section~\ref{subsection:consolidate})} slices and findings for faster iteration and community sharing. 
    \RGabbrv{} users can organize slices into a \testbench\ that can be versioned and shared, allowing a community of users to collaboratively build benchmarks and track progress.
    For reporting, \RGabbrv{} provides standard and custom robustness reports that can be auto-generated from testbenches and included in paper appendices.
\end{enumerate}

\tightpara{}To demonstrate how this process benefits practitioners, we outline how $3$ users with varying expertise can evaluate a natural language inference (NLI) model using \RGabbrv. 
Novice users (Section~\ref{personas:novice}) can rely on predefined testbenches for direct evaluation. 
Intermediate users (Section~\ref{personas:intermediate}) can create new slices using \slicebuilders\ available in \RGabbrv{}, and then construct their own testbenches. 
Finally, advanced users (Section~\ref{personas:expert}) can use their expertise to add custom slices. 
All of these users can generate a shareable Robustness Report (Figure~\ref{fig:robustness-report}).

\para{}We validate the Contemplate $\rightarrow$ Create $\rightarrow$ Consolidate process using a $3$-hour case study (Section~\ref{sec:sentiment}) with Salesforce's commercial sentiment modeling team. 
The team's main goal was to measure the bias of their model (\emph{contemplate}). 
We tested their system on $172$ slices spanning $3$ evaluation idioms%
, finding performance degradation on $12$ slices of up to $18\%$ (\emph{create}). 
Finally, we generated a single testbench and robustness report for the team, summarizing these findings (\emph{consolidate}).
A post-study questionnaire found that the team considered \RGabbrv{} to be easy to use, and indicated that they are very likely to integrate \RGabbrv{} into their workflow.
 
\para{}\RG{} can be used to conduct new research analyses with ease. To validate this, we conduct the first study of academic and commercially available named entity linking (NEL) systems, as well as a study of the fine-grained performance of summarization models. 
\begin{enumerate}[leftmargin=*]
    \item {\bf Named Entity Linking (Section~\ref{sec:nel})} We compare commercial APIs from \microsoft, \google\ and \amazon\ to open-source systems \bootleg, \wat\ and \rel\ across $2$ benchmark datasets (\wiki, \aida). We find that commercial systems struggle to link rare or less popular entities, are sensitive to entity capitalization and often ignore contextual cues when making predictions. \microsoft\ outperforms other commercial systems, while \bootleg\ displays the most consistent performance across a variety of slices. On \aida, we find that a simple heuristic NEL method outperforms all commercial systems.
    \item {\bf Summarization (Section~\ref{sec:summarization}).} We propose and implement $5$ subpopulations that capture summary abstractiveness, content distillation, information dispersion~\citep{grusky-etal-2018-newsroom}, positional bias, and information reordering~\citep{kedzie-etal-2018-content}. We compare $7$ models on the CNN-DailyMail dataset across these subpopulations. All models struggle on summaries that discard content, require higher amounts of abstraction or contain more entities. Surprisingly, models with very different prediction mechanisms make similar errors, suggesting that existing metrics are unable to capture meaningful performance differences.
\end{enumerate}

\tightpara{}\RG{} continues to be under active development, and we welcome feedback and suggestions from the community.

%% file: source/motivation_challenges.tex
\section{Current Evaluation Challenges}
We describe the problem of evaluating machine learning models, motivate a shift towards continual evaluation, lay out $3$ challenges today in making this shift, and situate this in the context of existing tools and work.

\subsection{Model Evaluation}
Generally, validation of a trained model for a task consists of evaluating the model on a set of examples that are drawn from the training distribution~\citep{Bishop2006PatternRA}.
Assuming identical train and test distributions (i.i.d. data), validation performance estimates the model's performance at test time.%

\para{}In practice, the train and test distributions can be {different}~\citep{Taori2020MeasuringRT}. 
This distributional shift is a natural consequence of changing real-world conditions and evolving expectations of the model's capabilities. %
For instance, a model that detects entities in news articles will become outdated as new entities emerge over time.
Standard validation overestimates true performance in this case, since it does not preempt performance degradation due to changing conditions.
Researchers and practitioners in these circumstances often rely on intuition and an understanding of their domain to create evaluations and perform model selection.

\para{}Recent work suggests that models often exploit spurious correlations when making predictions~\citep{mccoy2019right} and are not robust when evaluation moves beyond i.i.d. data~\citep{hendrycks2019using}. 
This lack of robustness makes models susceptible to failure under even the slightest distributional shifts~\citep{Miller2020TheEO}, or when deployed~\citep{Chloe2018Microsoft}. 
Systematic and continual evaluation is necessary to understand the model's limitations, and as we discuss next, standard evaluation practices often fall short.

\subsection{Towards Continual Evaluation}
We view evaluation as a continual process from the practitioner's perspective. In practice, constant re-evaluation is necessary in order to assess if a model should continue to be used in light of new information about its limitations. %
By contrast, traditional evaluation addresses challenges that relate to generating a static artifact of the model's performance (e.g., computing an aggregate measure of performance on a test set~\citep{Bishop2006PatternRA} or more fine-grained measures using a suite of tests~\citep{checklist:acl20}. 

\para{}Prior work on the construction of evolving benchmarks~\citep{dynabench} introduced dynamic evaluation, allowing a community of practitioners to collaboratively build challenging benchmarks. We focus here on the individual's perspective, and how to equip them with tools that support the continual evaluation paradigm. This raises a fresh set of challenges that are not traditionally addressed by standard evaluation. 

\para{}Next, we identify three of these challenges---the paradox of choice, idiomatic lock-in and workflow fragmentation---and highlight how existing tools and research fall short of addressing them. %

\subsection{Challenge 1: The Paradox of Choice}
\label{challenge:paradox_of_choice}
\para{Ideal.} Given a practitioner's task, needs, constraints and prior knowledge, give them guidance on what evaluation to run next.

\para{Challenges.} 
Evaluation is a complex, unstructured process for practitioners, since it can be confusing to choose what evaluation to run next. 
These decisions are frequent in continual evaluation. 
Here, practitioners accumulate an understanding of their model's limitations and manage changing needs, which should (ideally) guide future evaluation decisions. 
The goal is to help practitioners answer questions like: "Should I test for gender bias next?" or "Should I analyze generalization to longer inputs?". %

\para{}This aspect of evaluation remains understudied, because the focus has remained on prescribing and using a particular form of evaluation (e.g. inspect performance on perturbed examples).
Existing tools such as CheckList~\citep{checklist:acl20} and TextAttack~\citep{morris2020textattack} provide significant support on how to write code for particular evaluations, but give little guidance on what a user should run next. Existing research has studied questions related to the theory of generalization~\citep{hendrycks2020many} but very little is known about how to systematically evaluate models that will encounter distributional shift.

\para{}While there are no easy answers to these questions, we initiate a study of how practitioners can systematically make these decisions (Section~\ref{subsection:contemplate}), by identifying key decision variables such as their task, evaluation needs, resource constraints and history of evaluations.
    \begin{table*}[t!!]
\scriptsize
    \centering
    \begin{tabular}{lll}
    \toprule
        {\bf Evaluation Idiom} & {\bf Tools Available}  & {\bf Research Literature (focusing on NLI)}\\
        \midrule
        Subpopulations & Snorkel~\citep{ratner2017snorkel}, &Hard/easy sets~\citep{gururangan2018annotation}\\&Errudite~\citep{Wu2019ErruditeSR}& Compositional-sensitivity~\citep{nie2019analyzing}\\
        \cmidrule{2-3}
        \multirow{3}{*}{Transformations} & NLPAug~\citep{nlpaug}& Counterfactuals~\citep{kaushik2019learning}, Stress test~\citep{naik2018stress}, \\
        &&Bias factors~\citep{sanchez-etal-2018-behavior}, Verb veridicality~\citep{ross2019well}\\
        \cmidrule{2-3}
        \multirow{3}{*}{Attacks} & TextAttack~\citep{morris2020textattack}, &Universal Adversarial Triggers~\citep{wallace2019universal}, \\&OpenAttack~\citep{zeng2020openattack}&Adversarial perturbations~\citep{glockner2018breaking},\\&Dynabench~\citep{dynabench}&  ANLI~\citep{nie2019adversarial}\\
        \cmidrule{2-3}
        \multirow{2}{*}{Evaluation Sets} &  SuperGLUE diagnostic sets & FraCaS~\citep{cooper1994using}, RTE~\citep{dagan2005pascal}, SICK~\citep{marelli-etal-2014-semeval},\\
        &\citep{wang2019superglue}&SNLI~\citep{bowman2015large}, MNLI~\citep{N18-1101}, \\ &Checklist~\citep{checklist:acl20}&HANS~\citep{mccoy2019right}, Quantified NLI~\citep{geiger2018stress}, \\
         & & MPE~\citep{lai-etal-2017-natural}, EQUATE~\citep{ravichander-etal-2019-equate}, DNC~\citep{poliak-etal-2018-collecting-diverse}, \\&&ImpPres~\citep{jeretic2020natural},  Systematicity~\citep{yanaka2020neural} \\
         &&ConjNLI~\citep{saha2020conjnli}, SherLIiC~\citep{schmitt-schutze-2019-sherliic}\\
        \bottomrule
    \end{tabular}
    \caption{Tools and literature on robustness for NLP, with a focus on NLI as a case study. Some tools support multiple types of evaluations, for example, TextAttack supports both augmentations and attacks. For additional related work, see Section \ref{sec:related}.}
    \label{tab:tools}
\end{table*}

\subsection{Challenge 2: Idiomatic Lock-In}
\label{challenge:idiomatic_lock_in}
\para{Ideal.} Equip the practitioner with flexible tools to create and utilize evaluation examples that are best suited to the evaluation they want to run. %

\para{Challenges.} Once developers decide what they want to evaluate, they can suffer from lock-in to a particular \emph{idiom} of evaluation after they adopt a tool.  
Our analysis suggests that most tools and research today serve a subset of $4$ evaluation idioms:
\noindent
\begin{enumerate}[leftmargin=*]
    \item {\bf Subpopulations.} Identifying subpopulations of a dataset where the model may perform poorly. 
    
    {\it Example}: short reviews (< 50 words) in the IMDB review dataset. %
    
    \item {\bf Transformations.} Perturbing data to check that the model responds correctly to changes.

    {\it Example}: substituting words with their synonyms in the IMDB review dataset.%
    
    \item {\bf Attacks.} Perturbing data adversarially to exploit weaknesses in a model.

    {\it Example}: adding the word ``aaaabbbb" to the end of reviews makes the model misclassify. %
    
    \item {\bf Evaluation Sets.} Using existing datasets or authoring examples to test generalization and perform targeted evaluation.

    {\it Example}: authoring new movie reviews in the style of a newspaper columnist.
\end{enumerate}
These idioms are not exhaustive, but shed light on how evaluation is typically conducted. 
In Table~\ref{tab:tools}, we use this categorization to summarize the tools and research available today for the natural language inference (NLI) task. 
As an example, TextAttack~\citep{morris2020textattack} users can perform attacks, while CheckList~\citep{checklist:acl20} users author examples using templates, but cannot perform attacks. 

\para{}Tools vary in whether they provide scaffolding to let users build on new evaluation ideas easily. Tools often provide excellent abstractions for particular idioms, (e.g., TextAttack~\citep{morris2020textattack} scaffolds users to easily write new adversarial attacks). However, no tool that we are aware of addresses this more broadly for evaluation that cuts across idioms.

\para{}All of these limitations can make it difficult for practitioners, who are forced to glue together a combination of tools. Each tool meets different developer needs, and has its own abstractions and organizing principles, which takes away time from users to inject their own creativity and expertise into the evaluation process.

\para{}We address these challenges with \RG{} (Section~\ref{subsection:create}), which uses an open-interface design to support all $4$ evaluation idioms, and provides a simple workflow to scaffold users.

\subsection{Challenge 3: Workflow Fragmentation}
\label{challenge:workflow_fragmentation}
\para{Ideal.} Enable practitioners to store, version and share evaluation data, communicate findings and collaborate to take advantage of others' work. %

\para{Challenges.}
As practitioners evaluate, they need to keep track of progress and communicate results. Evaluation tools today let users save their progress, but provide no support for semantic versioning~\citep{preston2013semantic} and sharing findings.  This is made more difficult when trying to consolidate evaluations and results across multiple tools. General-purpose data storage solutions solve this problem, but require significant user effort to customize.%

\para{}Reporting findings can be difficult since there is no consensus on how to report when performing evaluation across multiple idioms. Attempts at standardized reporting suggest guidelines for what to report~\citep{mitchell2019model}, but are free-form, leaving the responsibility of deciding what to report to the user. 
\begin{table}
\small
    \centering
    \begin{tabular}{lrr}
    \toprule
        
       \multicolumn{2}{c}{ \# Model Cards } & \% of Models\\
        \midrule
         Total & 2133 & 64.6\% \\
         Non-empty & 922 & 27.9\% \\
         Any evaluation info & 197 & 6.0\% \\
         \midrule
         \# Models & 3301 & 100.0\%\\
         \bottomrule
    \end{tabular}
    \caption{\small Prevalence of evaluation information in model cards on the HuggingFace Model Hub (\url{huggingface.co/models}).}
    \label{tab:hfmodels}
\end{table}

\para{}To study whether existing tools incentivize reporting, we scraped model cards~\citep{mitchell2019model} for all available Huggingface models~\citep{wolf2020huggingfaces} (as of 09/22/2020). Model cards are free-form templates for reporting that contain an entry for ``Evaluation" or ``Results", but leave the decision of what to report to the user. Huggingface provides tools for users to create model cards when submitting models to their model hub.

\para{}Our findings are summarized in Table~\ref{tab:hfmodels}. Only a small fraction ($6.0\%$) of models carry model cards with any evaluation information. Qualitatively, we found low consistency in how users report findings, even for models trained on the same task. This suggests that it remains difficult for users to report evaluation information consistently and easily.

\para{}In Section~\ref{subsection:consolidate}, we describe the support that \RG{} provides for versioning evaluations in testbenches, and easily exporting and reporting findings with Robustness Reports. %

\begin{figure*}[!t]
    \centering
    \includegraphics[width=\textwidth]{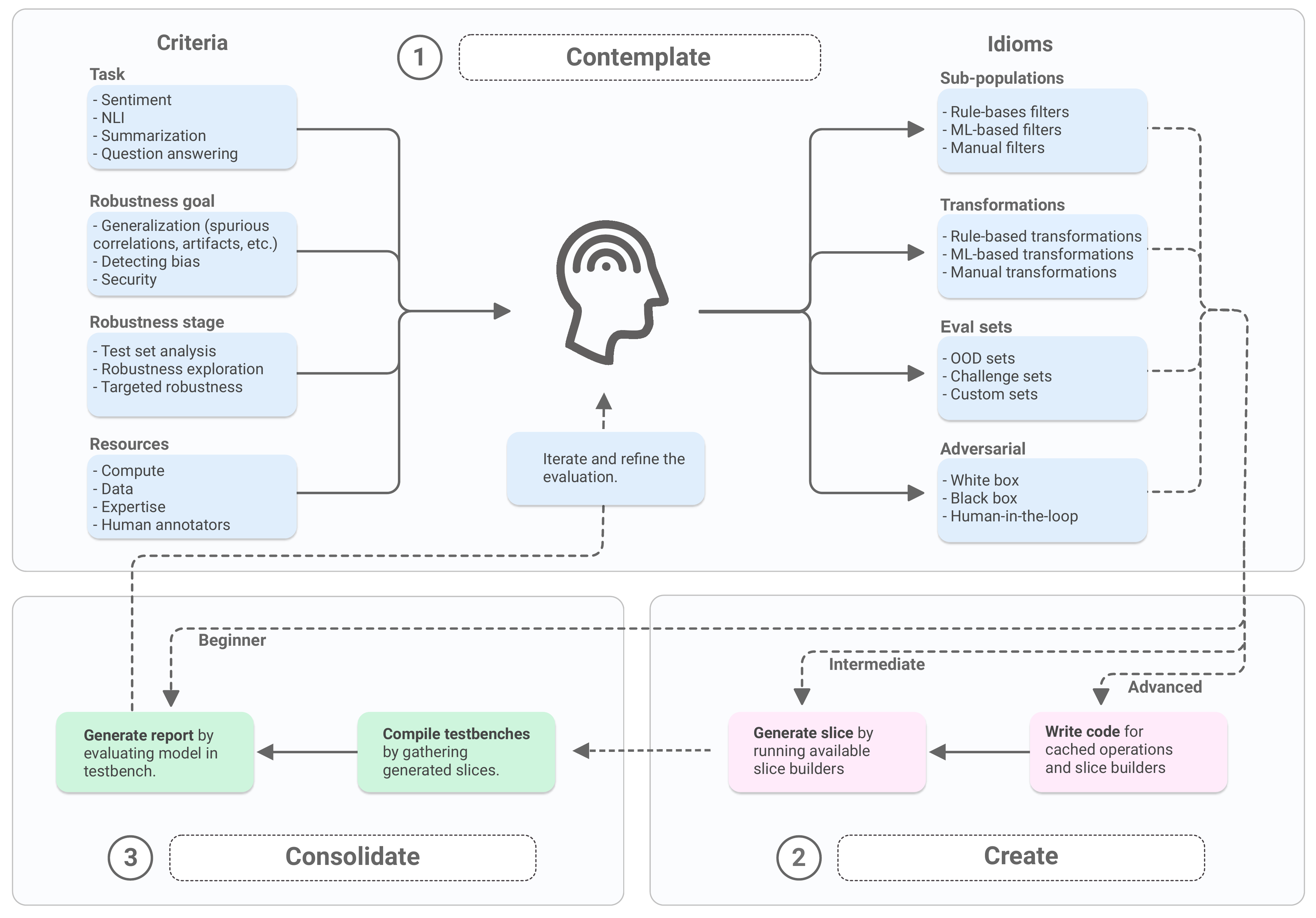}
    \caption{Illustration of the Contemplate $\rightarrow$ Create $\rightarrow$ Consolidate loop (Section~\ref{sec:3cs}).}
    \label{fig:process}
\end{figure*}

%% file: source/3cs.tex
\section{Continual Evaluation Workflow}
\label{sec:3cs}

To address the challenges we highlighted in the previous section, we propose the Contemplate $\rightarrow$ Create $\rightarrow$ Consolidate loop for performing continual evaluation. 
In this framework, practitioners will
\begin{enumerate}[leftmargin=*]
    \item {\bf Contemplate} (Section~\ref{subsection:contemplate}) what evaluation to run next, using guidance on key decision variables, %
    \item {\bf Create} (Section~\ref{subsection:create}) slices of data for evaluation using \RG,
    \item {\bf Consolidate} (Section~\ref{subsection:consolidate}) findings using the testbench and reports in \RG.
\end{enumerate}
Figure~\ref{fig:process} illustrates this continual evaluation loop, which we describe in more detail next.

%% file: source/contemplate.tex
\subsection{Contemplate: Navigating the Evaluation Landscape}
\label{subsection:contemplate}
As we highlighted in Section~\ref{challenge:paradox_of_choice} (Paradox of Choice), practitioners may find it difficult to choose the appropriate evaluation among the large number of possibilities. %
We provide guidance to practitioners by focusing on key decision variables: the task, the evaluation goal, the resource constraints, and the history of prior evaluations. We connect these variables to decisions about which evaluation idiom and particular evaluations may be most appropriate. 
We emphasize that there is no ``silver bullet'' here, and our goal is to initiate research in how to make these decisions systematically.  %

\para{}Figure~\ref{fig:process} visualizes and enumerates these decision criteria (top-left), embeds them in our evaluation loop, and highlights the actions available to the user in the form of which evaluation idiom and specific evaluation to choose (top-right). We describe the decision criteria below.%

\para{Task.} We consider scenarios when the practitioner's task can suggest evaluations that are already known or easily available:

\begin{itemize}[leftmargin=*]
    \item {\bf Existing research.} 
    Prior work serves as a good starting point for evaluations that models commonly succeed against, as well as those where models typically fail (e.g., for NLI, it is well-known that examples with negation~\citep{naik2018stress}  are difficult for models to classify).
    \item {\bf Datasets.} 
    Tasks that are well-studied have evaluation sets that are publicly available (e.g., MNLI~\citep{N18-1101} and HANS~\citep{mccoy2019right} for NLI). These serve as a useful starting point for evaluation, although users should be aware of the danger of overfitting to these evaluation sets~\citep{Dwork2015TheRH}. 
    \item {\bf Input/output structure.} 
    The structure of the task may constrain the types of evaluations that may be performed. For example, subpopulations based on lexical overlap may only be applied when the task is a function of two or more inputs (e.g., natural language inference accepts as input a premise and hypothesis). Prior research on similarly structured tasks can provide inspiration on a new or understudied task.
 \end{itemize}

\para{Evaluation goals.}  We consider $3$ broad evaluation goals: testing generalization (spurious correlations, sensitivity to noise, distributional artifacts), detecting bias (gender bias, dependence on sensitive attributes), and ensuring security (vulnerability to malicious users). The practitioner's interest in these goals should influence the evaluations they choose. 

\begin{itemize}[leftmargin=*]
    \item {\bf Generalization.} Predefined out-of-distribution data splits may be used to evaluate a model's ability to generalize outside of the specific dataset set on which it was trained~\citep{gardner2020evaluating,koh2020wilds}. Challenge datasets (e.g., HANS~\citep{McCoy2019RightFT}), can identify a model's ability to overcome spurious correlations in the training set (e.g., lexical overlap). Similarly, subpopulations can be constructed to leverage existing examples in a model to test particular generalization capabilities. Transformations such as paraphrasing may be used to augment the dataset with examples with differing surface-level features to test the model's reliance on artifacts in the training set. 
    
    \item {\bf Detecting bias.} Depending on the task,  evaluation sets may be available to test for a model's bias with respect to particular protected attributes (e.g., gender bias in coreference in the case of Winogender~\citep{rudinger2018gender} and Winobias~\citep{zhao2018gender}). If no existing datasets exist, they may be synthesized by performing hand-crafted transformations with respect to particular protected attributes~\citep{10.1145/3375627.3375865} or subpopulations that contain particular groups considered.
    
    \item {\bf Security.}  A user might be interested in security and understanding their system's vulnerabilities, for example, a spammer may try to use adversarial attacks to bypass a spam email filter~\citep{biggio2013evasion}. Towards this end, the user should focus their evaluations on adversarial attacks.
\end{itemize}

\paragraph{Resource constraints.} Constraints are central to determining the evaluations feasible for a practitioner.      

\begin{itemize}[leftmargin=*]
    \item {\bf Compute.} If compute is limited (e.g., no GPUs are available), subpopulations may be most appropriate since they can reuse predictions while attacks should be avoided since they can be extremely compute-intensive.
    \item {\bf Data.} Access to data can be a bottleneck that dictates what evaluations are possible. Some tasks may require the use of proprietary or protected data (e.g., clinical notes in hospitals, or customer data in a company, making procurement and use more difficult). Transformations applied to existing data, such as with generative modeling, can be valuable in narrowing the data gap in this case.
    \item {\bf Human resources.} Some evaluation strategies require a large amount of manual effort (e.g., creating custom evaluation sets).  Evaluation strategies that require constructing hand-crafted rules (e.g., subpopulations), may also be time consuming. Standard transformations (e.g., paraphrasing), that augment existing datasets may help alleviate these efforts, or automated approaches to creating synthetic datasets (e.g., few-shot generation using GPT-3 \citep{brown2020language}), may be preferred.
    \item {\bf Expertise.} A user's expertise will determine whether they are able to create custom evaluations versus relying on existing ones. Domain expertise may be required to author custom evaluation sets or write custom rules for generating subpopulations. Technical expertise may be needed to write customized code for certain types of robustness tests (e.g. adversarial attacks), and should be sought if required.
\end{itemize}

\paragraph{Prior evaluations.} The history of prior evaluations and the stage of robustness testing will also influence the choice of the next evaluation to perform. We describe $4$ evaluation strategies that practitioners can use to guide continual evaluation efforts.

\begin{itemize}[leftmargin=*]
    \item {\bf Easy $\rightarrow$ Hard.} Initial evaluations might focus on simple tests such as robustness to standard transformations (e.g., synonym substitution). Models shown to be robust against these simpler tests might then be tested on harder challenge sets or adversarial attacks. 
    \item {\bf Coarse $\rightarrow$ Fine.} Early evaluation should typically focus on coarse evaluations with large slices of data (e.g., performance on long vs. short inputs). Later stages of evaluation should drill-down into fine-grained slices of relevance to model deployment (e.g., queries about the Beatles in a question-answering system).
    \item {\bf Explore $\rightarrow$ Exploit.} Early evaluation stages are more exploratory, as users sift through a large number of slices to search for weaknesses. Over time, it becomes more clear where a model is more or less performant, and later evaluation stages can exploit this knowledge to develop a more fine-grained understanding of performance.
    \item {\bf Generic $\rightarrow$ Targeted.} Initial evaluations can draw on prior knowledge and community know-how of common evaluations. As evaluation proceeds, focus shifts to developing new evaluations that are most appropriate to the user's goal of deploying their model.
\end{itemize}
As evaluation proceeds, users should consider keeping prior evaluations as a form of regression testing~\citep{Wahl1999AnOO}. Much like in software, changes to the model should not degrade performance on slices where the model previously performed well. 

%% file: source/create.tex
\subsection{Create: \RG{}}
\label{subsection:create}
\para{}As highlighted in Section~\ref{challenge:idiomatic_lock_in} (Idiomatic Lock-In), practitioners can get locked into a single tool that supports only a few evaluation idioms. 
We introduce \RG{} (\RGabbrv), a toolkit that enables broad evaluation across multiple idioms. 
Figure~\ref{fig:rg-system-design} provides a visual depiction of the abstractions in \RGabbrv{} while Python examples for \RGabbrv{} are in Tables~\ref{tab:code-cachedops}, \ref{tab:code-slicebuilders} and \ref{tab:code-reporting} of the appendix. At a high level, \RGabbrv{} breaks robustness testing into a two-stage workflow:
\begin{enumerate}[leftmargin=*]
    \item {\bf Caching information.} First, practitioners typically perform a set of common pre-processing operations (e.g., tokenization, lemmatization) and compute useful side information for each example (e.g., entity disambiguation, coreference resolution, semantic parsing) using external knowledge sources and models, which they cache for future analysis. 
    
     A large part of practitioner effort goes into generating this side information---which can be expensive to compute---and into standardizing it to a format that is convenient for downstream analysis. This layer of complexity can make it difficult for them to share their evaluation with others.
    
    \emph{RG Support.} \cachedop~is an abstraction in \RGabbrv~to derive useful information or generate side information for each example in a dataset by (i) letting users run common operations easily and caching the outputs of these operations (e.g., running the spaCy pipeline~\citep{spacy}); (ii) storing this information alongside the associated example so that it can be accessed conveniently; (iii) providing a simple abstraction for users to write their own operations.
    
    \item {\bf Building slices.} Second, practitioners use the examples' inputs and any available cached information to build \textit{slices}, which are collections of examples for evaluation based on any of the $4$ evaluation idioms. 
    
    \emph{RG Support.} \slicebuilder~is an abstraction to retrieve available information for an example and create slices of data from them by (i) providing retrieval methods to access inputs and cached information conveniently when writing custom code to build slices; (ii) providing specialized abstractions for specific evaluation idioms: transformations, attacks and subpopulations.
\end{enumerate}
This breakdown naturally separates the process of gathering useful information from the nuts and bolts of using that information to build slices. Table~\ref{tab:rg-slicebuilders} contains examples of \cachedops~and \slicebuilders~ that will be available in \RG.  %

 \para{}\RG{} relies on a common data interface provided by the datasets library from HuggingFace \citep{wolf2020huggingfaces}, which is backed by Apache Arrow~\citep{arrow}. This ensures that all operations in \RG{} interoperate with HuggingFace models, and can be exported easily.

%% file: source/consolidate.tex
\subsection{Consolidate: Share Testbenches and Findings}
\label{subsection:consolidate}
As highlighted in Section~\ref{challenge:workflow_fragmentation} (Workflow Fragmentation), users can find themselves consolidating evaluation results across several tools and evaluation idioms. 
\RG{} addresses this fragmentation by providing users a \testbench\ abstraction. 
Using this, users can assemble and version a collection of slices, which represents a suite of evaluations. 
\RG{} tracks the provenance of these slices, making it possible to identify (i) the data source that the slice originated from; (ii) the sequence of \slicebuilders~by which a slice was constructed. 
This makes it possible for another user to reproduce or redo analysis in a collaboration, through sharing of a \testbench.

\para{}\RG{} also provides a general-purpose tool for creating \emph{Robustness Reports} for any model on a \testbench. Users can also use Robustness Reports on their own, allowing them to generate reports for evaluations that are not performed in \RGabbrv{}. 

\para{}To incentivize standardization in reporting, \RGabbrv{} includes \emph{Standard Reports} for several tasks. 
The Standard Report is comprehensive, static and is backed by a \testbench\ that contains slices from all evaluation idioms. It can either be generated in a PDF or \LaTeX~format to be added to the appendix of a paper\footnote{See Figure~\ref{fig:latex-report} in the appendix.}. Reports reduce user burden in communicating findings, and make it easier to standardize reporting in the community.
In the future, \RG{} will also include an interactive tool for generating reports that allows users to pick and choose slices of interest based on their robustness goals and constraints.

%% file: source/personas.tex
\section{User Personas in \RG{}}

In this section, we discuss how users with varying expertise can use \RG{} to perform continual evaluation and robustness testing.  We describe user personas at $3$ skill levels---beginner, intermediate, and advanced---and explain a possible path through the Contemplate $\rightarrow$ Create $\rightarrow$ Consolidate process for each of them. In every case, we assume that the user's goal is to analyze the performance of an NLI model. Figure~\ref{fig:process} illustrates how these user personas can be situated into this workflow.

%% file: source/novice.tex
\subsection{Scenario \rom{1}: Beginning User}
\label{personas:novice}

\para{Contemplate.}
The user's goal is to perform exploratory robustness testing for the NLI task. Because the user is new to NLP and robustness testing, they lack the knowledge to choose specific slices or write custom slices. Therefore they decide to run the Standard Report for NLI.

\para{Create.}
The user is able to create the report with a few clicks in the \RGabbrv{} interface. They select ``Standard Report'', ``Ternary Natural Language Inference'' (task), ``SNLI'' (dataset),  ``BERT-Base'' (model), and click ``Generate Report''. 

\para{Consolidate.}
The Standard Report, shown in Figure~\ref{fig:robustness-report} provides a detailed snapshot of various robustness tests for NLI models. 
The tests may include Subpopulations (e.g., {\sc HasNegation}, {\sc LexicalOverlap}), Transformations (e.g., {\sc SynonymAug}, {\sc KeyboardAug})~\citep{nlpaug}, Attacks ({\sc TextAttack})~\citep{morris2020textattack, Garg2020BAEBA}, and Evaluation Sets~\citep{bowman2015large}. %
The user gleans several initial insights from this report. 
For example, they see that the model is vulnerable to common typing mistakes due to low accuracy on the {\sc KeyboardAug} slice; the predicted class distribution column further reveals that this noise causes the model to predict \texttt{contradiction} significantly more frequently than \texttt{entailment} or \texttt{neutral}. 
The user is able to easily share the generated PDF of this report with their colleagues, with whom they can iterate on additional robustness tests for misspellings.

%% file: source/intermediate.tex
\subsection{Scenario \rom{2}: Intermediate User}
\label{personas:intermediate}

\para{Contemplate.} 
This user is interested in exploring gender bias in NLI models. Specifically they would like to test cases where specific gendered pronouns are present in the premise or hypothesis.  
They are willing to write minimal code to instantiate existing \slicebuilder\ classes with custom parameters but do not want to write the code from scratch. 
Therefore they decide to create slices using built-in subpopulation \slicebuilders.

\para{Create.}
The user applies the existing {\sc HasPhrase} class in order to create subpopulations with female pronouns in the hypothesis:

\begin{minipage}{0.95\linewidth}
\begin{lstlisting}[language=Python, numbers=none, style=inlinestyle]
     subpopulations = HasPhrase(['her', 'she']) # instantiate
     slices = subpopulations(snli, ['hypothesis']) # apply to data
\end{lstlisting}
\end{minipage}

\para{Consolidate.} The user generates a report for immediate analysis and makes the \testbench\ available on GitHub in order to collaborate with the broader community.  

%% file: source/expert.tex
\subsection{Scenario \rom{3}: Advanced User}
\label{personas:expert}

\para{Contemplate.}  This user is interested in performing robustness tests for spurious correlations in NLI related to surface-level similarities between premise and hypothesis. 
They are particularly interested in evaluating whether models rely on the premise and hypothesis being of similar length in order to detect entailment. 
As they are performing a novel analysis, they plan on writing custom logic to create the appropriate slices. They consider two types of slices: subpopulations and transformations, as described below.

\para{Create.} 
\para{}The user utilizes the existing {\sc ScoreSubpopulation} class, which constructs subpopulations using arbitrary scoring functions. They create a custom scoring function \texttt{len\_diff}, which returns the absolute difference in length between the hypothesis and premise, and then create a  \slicebuilder\ for the subpopulation of examples that score in the top $10\%$ as follows:

\begin{minipage}{0.95\linewidth}
\begin{lstlisting}[language=Python, numbers=none, style=inlinestyle]
     s = ScoreSubpopulation(intervals=[('90%','100%')], score_fn=len_diff)
\end{lstlisting}
\end{minipage}

\para{}The user also utilizes existing  \slicebuilders\, such as the {\sc LexicalOverlap} class, which creates subpopulations based on the lexical overlap between premise and hypothesis. Additionally, they transform the dataset using classes such as {\sc EasyDataAugmentation}~\citep{wei2019eda}. They can then compose this transformation with the custom \slicebuilder\ described earlier to create a larger evaluation set.

\para{Consolidate.} The user generates a report for immediate analysis, and also generates an appendix for a paper to share results with the research community. They make their code and testbench available on GitHub so that others may reuse and refine their approach.

\begin{figure*}[t]
    \centering
    \includegraphics[width=\textwidth]{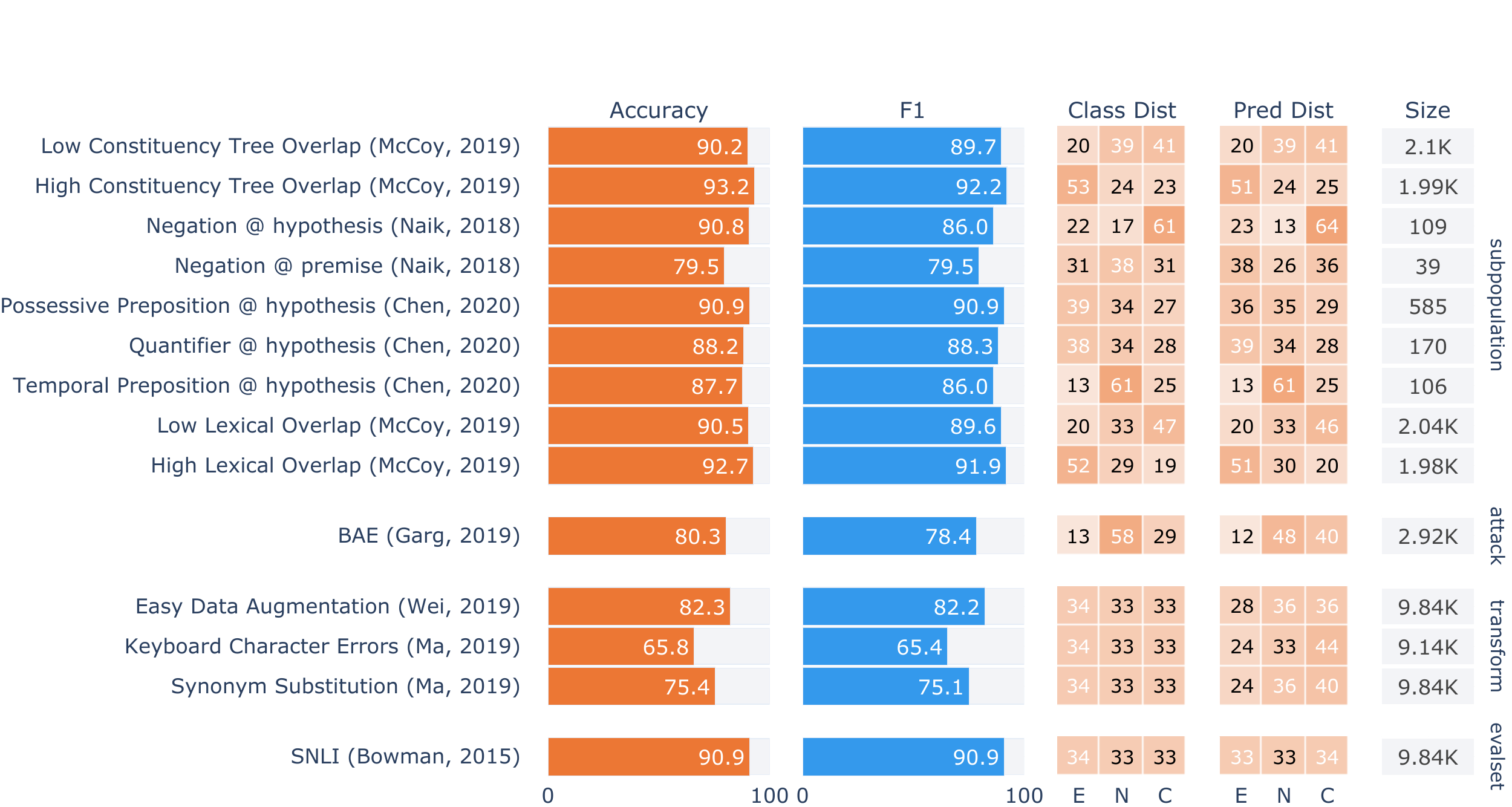}
    \caption{Robustness Report for Natural Language Inference using {\tt bert-base} on SNLI.} %
    \label{fig:robustness-report}
\end{figure*}

%% file: source/experiments/commercial.tex
\subsection{Commercial Sentiment Analysis Case Study}
\label{sec:sentiment}

We validate the Contemplate $\rightarrow$ Create $\rightarrow$ Consolidate workflow with \RG{} through a real-world case study. 
We conducted a 3-hour long virtual case study with a member of the team that built the Einstein sentiment system which is part of Salesforce's cloud offerings.\footnote{\url{https://einstein.ai/products/community-sentiment}}

\para{Pre-study questionnaire.} Our pre-study questionnaire elicited information on the team's task (e.g., sentiment modeling, question answering, etc.), what metrics they use for evaluation (e.g., accuracy, F1, etc.), how they evaluate robustness (e.g., standard validation/testing, out-of-distribution data, bias testing, attacks) and what their evaluation goals are (e.g., security, generalization). Their responses suggest that their evaluations were mainly on a proprietary validation set that included some out-of-distribution data, and their main interest was in understanding the potential bias for identity groups.

\para{}We also asked them to rate on a Likert scale ($1-5$), whether they would
``like to evaluate the robustness of [their] model more thoroughly than [they] do today.'' (agreement $4/5$) and ``would benefit from having a library that gives [them] the tools to evaluate the robustness of [their] models.'' (agreement $5/5$).
The format and other details about the questionnaire are in Appendix~\ref{app:study}. 

\para{\bf Study.} The study took place during the COVID-19 pandemic and was conducted virtually with the user from the sentiment team. Due to difficulties in guiding them virtually, one of the authors shared their screen and conducted all the experimentation throughout the 3-hour period. 

\para{}We followed the Contemplate $\rightarrow$ Create $\rightarrow$ Consolidate loop, and we highlight the key steps that we took through the study period.
\begin{itemize}[leftmargin=*]
    \item {\bf Contemplate (1).} We first identified resource constraints---the user provided us with their evaluation data, and gave us black-box access to their model\footnote{We could not access the model directly, but could give them examples to fetch predictions.}. We used a CPU-only MacBook Pro for all computation. Since the team had previously analyzed the sensitivity of their model to mentions of race/gender/religion/ethnicity, we decided to first verify performance on subpopulations of their dataset with identity-sensitive words.
    \item {\bf Create (1).} We constructed slices for evaluating performance using a \slicebuilder~that searched for identity words. We found no degradations compared to the average performance of the model on nine identity-sensitive words.
    \item {\bf Contemplate (2).} Next, after discussion with the user, we considered whether the model could have performance disparities along different topics. %
    \item {\bf Create (2).} We next evaluated the model on subpopulations that contained topic-specific words. We found that the model did poorly on some topics, with performance degradations of up to $18\%$.
    
    \item {\bf Contemplate (3).} Next, we set out to understand whether the model was robust to input perturbations. The user highlighted instances of input noise that they wanted to gather more information on.
    \item {\bf Create (3).} We used $4$ different transformations for simulating typing errors and paraphrasing text, and found that performance degraded by $6\%$.
    
    \item {\bf Contemplate (3).} Lastly, they wanted to investigate whether the model was robust to larger distributional shifts in the inputs. 
    \item {\bf Create (3).} We downloaded and used an open-source sentiment dataset, and found that performance degraded by $5\%$.
    \item {\bf Consolidate (1).} We collated all the slices into a testbench, and generated a report to share with other members of their team.
    
\end{itemize}
We performed $3$ iterations of (Contemplate $\rightarrow$ Create), resetting the evaluation objective after each iteration, and using \RG{} to investigate them. Overall, we evaluated the system on $172$ different subpopulations, $1$ open-source evaluation set from the internet, and $4$ different transformations, all in the $3$-hour period. 
We observed a total of $12$ subpopulations where performance degraded significantly. 
This performance deterioration occurred under all $4$ types of transformations as well. 
Lastly, since we did not have access to the model for training, we made prescriptions for augmentation-based training to improve performance on examples where the model underperformed. 

\para{Post-study questionnaire.} We conducted a post-study questionnaire with the user, where we asked them to provide feedback on \RG{} and the overall study. We elicited feedback on ``how likely [they] were to incorporate \RG\ in [their] workflow" (very likely $5/5$), and the perceived ``ease of use of \RG" (high $5/5$). In feedback related to the utility of the $4$ evaluation idioms in \RG{}, they found subpopulations to be ``very insightful", and were enthusiastic about the ability to perform various evaluations in a single tool. Lastly, the robustness report gives information on how the team could make improvements and work towards adopting continual evaluation for their system.

%% file: source/experiments/experiments.tex
\section{Experimental Results using \RG{}}
\RG{} makes it easy for researchers and practitioners to perform novel analyses of existing tasks and models. 
To demonstrate this, we use \RG{} to investigate fine-grained performance on $2$ tasks---named entity linking (NEL) and text summarization.   
For NEL, we present the first fine-grained analysis of NEL across $3$ widely used commercial APIs, and $3$ state-of-the-art academic systems. For summarization, we analyze $7$ state-of-the-art models for text summarization trained on the CNN/DailyMail dataset. %

%% file: source/experiments/ned.tex
\subsection{NEL on Commercial APIs}
\label{sec:nel}
We analyze the fine-grained performance of commercial and state-of-the-art-systems for named entity linking (NEL). 
NEL is a fundamental component of both search and question-answering systems such as conversational assistants, and has a widespread impact on the performance of commercial technology. 
Given some text, the NEL task involves the identification of all entity mentions, and contextualized linking of these mentions to their corresponding Wikipedia entries, e.g., ``She drove a Lincoln to Lincoln" would link the first mention of Lincoln to \texttt{Lincoln\_Motor\_Company} and the second mention of Lincoln to \texttt{Lincoln,\_Nebraska}. Each identified mention (e.g., ``Lincoln") is typically mapped to a candidate list of Wikipedia entries (e.g., all ``Lincoln"-related Wikipedia entries) before disambiguation.
Our goal is to use \RG{} to understand where existing NEL systems fall short. 

\para{Systems.} We consider 3 commercially available NEL APIs: 
(i) \google\ Cloud Natural Language API\footnote{\url{https://cloud.google.com/natural-language}},
(ii) \microsoft\ Text Analytics API\footnote{\url{https://azure.microsoft.com/en-us/services/cognitive-services/text-analytics/}}, and 
(iii) \amazon\ Comprehend API\footnote{\url{https://aws.amazon.com/comprehend/}}\footnote{\amazon~only performs named entity recognition (NER) to identify mentions of named-entities in text, so we use it in conjunction with a simple string matching heuristic to resolve entity links.}. 
We compare them to $3$ state-of-the-art systems and a heuristic baseline: 
(i) \bootleg\ \citep{Orr2020BootlegCT}, a self-supervised system for NEL, 
(ii) \rel\ \citep{Hulst2020RELAE}, a system that combines existing state-of-the-art approaches, 
(iii) \wat\ \citep{Piccinno2014FromTT} an extension of the TAGME~\citep{Ferragina2010TAGMEOA} linker, and 
(iv) \pop, our simple heuristic baseline that picks the most popular entity among a set of candidate entities.

\para{Datasets.} We compare these methods on examples drawn from two datasets: 
(i) \wiki, which contains $100,000$ entity mentions across $37,492$ sentences from a $2019$ Wikipedia dataset, and 
(ii) \aida, the AIDA test-b dataset.

\para{Metrics.} For \wiki, we compare performance on recall\footnote{\wiki~is sparsely labeled and we do not report precision or F1 scores, which can be misleading.}. For \aida, we compare performance on Macro-F1.

\begin{figure*}[t!]
    \centering
    \includegraphics[width=\textwidth]{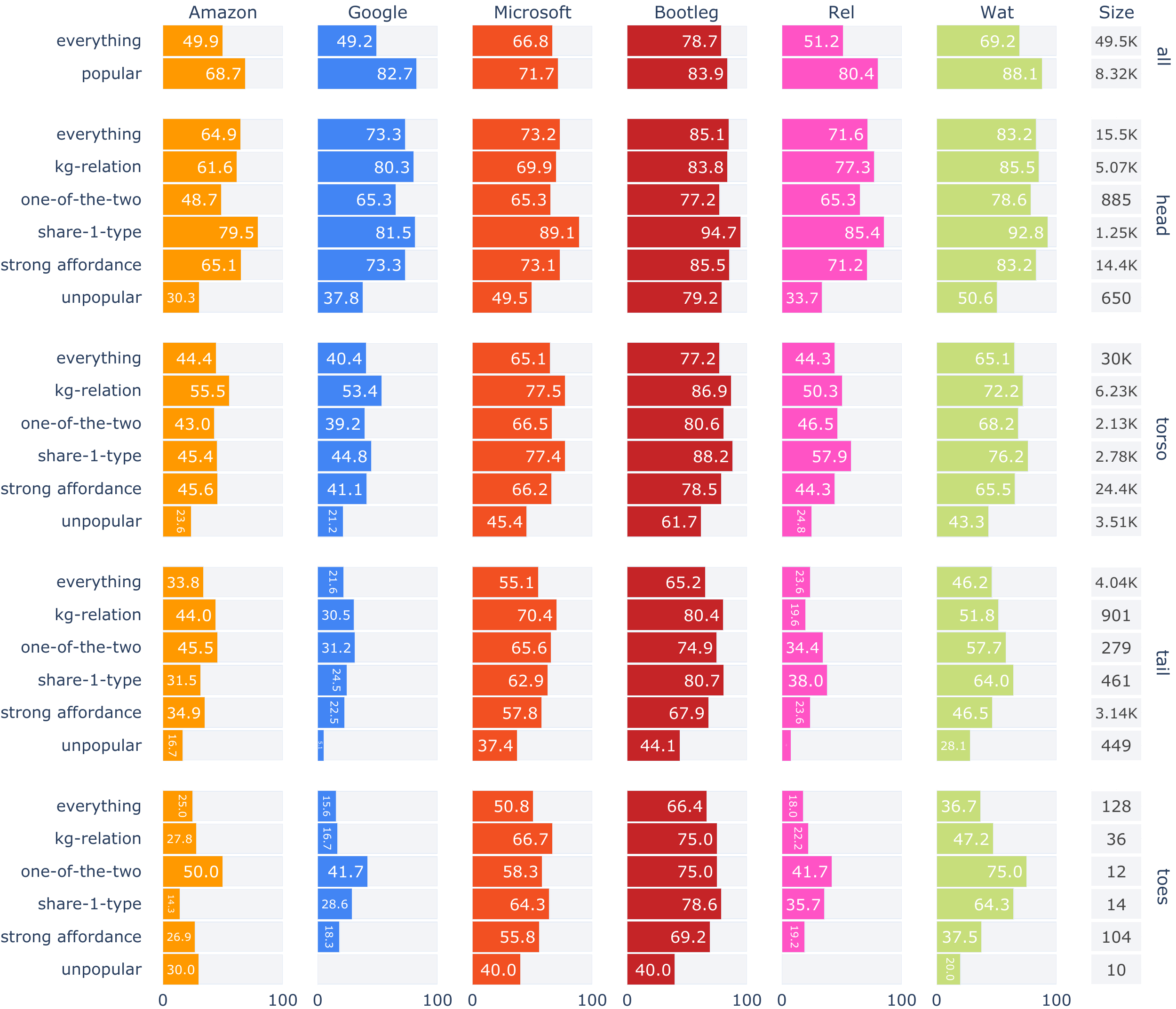}
    \caption{Robustness Report for NEL on Wikipedia. Performance reported using the Recall metric.}
    \label{fig:wiki-robustness-report}
\end{figure*}

\subsubsection{Analysis on \wiki}
\noindent{\bf Slices.} In line with~\cite{Orr2020BootlegCT}, we consider $4$ groups of slices---head, torso, tail and toe---that are based on the popularity of the entities being linked. Intuitively, head examples involve resolving popular entities that occur frequently in \wiki, torso examples have medium popularity while tail examples correspond to entities that are seen rarely. Toe entities are a subset of the tail that are almost never seen. We consider $5$ subpopulations from~\citep{Orr2020BootlegCT} within each group, 
\begin{itemize}[leftmargin=*]
    \item {\it kg-relation} contains examples where the entity being linked is related to another entity in the sentence. This serves as useful contextual information that can aid disambiguation.
    \item {\it one-of-the-two} contains examples where the gold entity is one of the two most popular candidates in the list of candidates, and both have similar popularity. These examples require careful use of context to disambiguate.
    \item {\it share-1-type} contains examples where the sentence contains $3$ consecutive entities that share the same type affordance. These type affordances can be used as contextual cues for disambiguation.
    \item {\it strong-affordance} contains examples where the sentence has words that are highly associated (as measured by tf-idf) with the gold entity's type(s). Again, these words can be used as contextual cues.
    \item {\it unpopular} contains examples where the gold entity is the second or less popular entity in the list of candidates, and the most popular entity is at least $5\times$ more popular than the second. These examples require the model to overlook popularity in favor of preferring a more uncommon entity.
\end{itemize}
Lastly, we also consider performance on \emph{popular} entities which correspond to examples where the entity mention corresponds to one of the top $800$ most popular entity mentions.

\para{Bootleg is best overall.} Overall, we find that \bootleg\ is the best-performing system, while \microsoft\ is the best-performing commercial system. \bootleg\ outperforms other systems by a wide margin, with a $12$ point gap to the next best system (\microsoft), while \microsoft\ in turn outperforms other commercial systems by more than $16$ points. 

\para{Performance degrades on rare entities.} For all systems, we find that performance on head slices is substantially better than performance on tail/toe slices. \bootleg\ is the most robust across the set of slices that we consider\footnote{We note that this may partly be a consequence of the set of slices we use, which are taken from \cite{Orr2020BootlegCT}.}. In particular, we note that \google\ and \amazon\ struggle on tail and torso entities, while \microsoft's performance degrades more gracefully. \google's model is particularly adept at popular entities where it outperforms \microsoft\ by more than $11$ points.

\subsubsection{Analysis on \aida}
For \aida, we compare performance on Macro-F1, since \aida\ provides a dense labeling of entities (and therefore computing precision is meaningful). Similar to \wiki, we find that \bootleg\ is the best-performing system overall on \aida, while \microsoft\ is the best-performing commercial system. 

\para{Sensitivity to capitalization.} Both \google\ and \amazon\ are sensitive to whether the entity mention is capitalized. \google's performance goes from $54.1$\% on sentences where all gold-labeled entities are capitalized to $38.2$\% on sentences where no gold-labeled entities are capitalized. Similarly, \microsoft\ degrades from $66.0$\% to $35.7$\% on these slices. This suggests that mention extraction in these models is quite sensitive to capitalization. In contrast, \amazon, \bootleg\ and \wat\ have stable performance, regardless of capitalization.

\para{Performance on topical entities.} Interestingly, all models appear to struggle on some topical slices (e.g., on the NFL slice), all models degrade significantly, with \bootleg\ outperforming other models by $20$+\%. Both \google\ and \microsoft\ display strong performance on some topics, (e.g., \google\ on alpine sports and \microsoft\ on skating).

\begin{figure*}[!t]
    \centering
    \includegraphics[width=\textwidth]{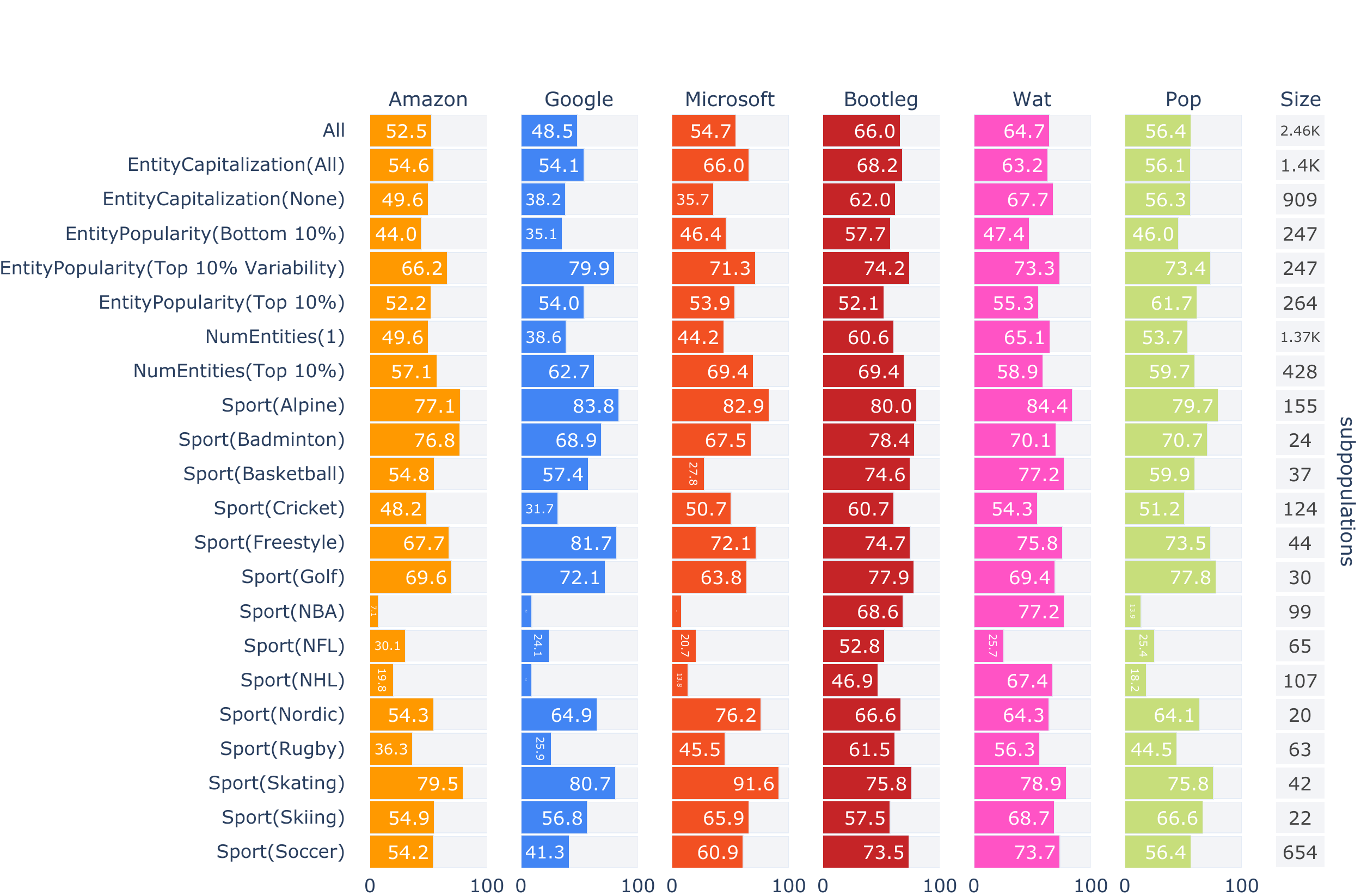}
    \caption{Robustness Report for NEL on AIDA. Performance reported using the Macro-F1 metric.}
    \label{fig:wiki-robustness-report}
\end{figure*}

\para{Popularity heuristic outperforms commercial systems.} Somewhat surprisingly, we find that {\sc Pop} outperforms all commercial systems by $1.7$ points. In fact, we note that the pattern of errors for {\sc Pop} is very similar to those of the commercial systems (e.g., performing poorly on NBA, NFL and NHL slices). This suggests that commercial systems sidestep the difficult problem of disambiguating ambiguous entities in favor of returning the more popular answer. Similar to \wiki\, \google\ performs best among commercial systems on examples that contain the most popular entities (top $10\%$ entity popularity).

\para{}Overall, our results suggest that state-of-the-art academic systems substantially outperform commercial APIs for NEL.

%% file: source/experiments/summarization.tex
\subsection{Summarization with State-of-the-Art Models}
\label{sec:summarization}
Next, we analyze the performance of state-of-the-art summarization models using \RG. We selected summarization as an instance of a  text-generation task to demonstrate the versatility of \RG{} for prediction tasks beyond classification or sequence labeling. For example, we show how slices can be computed based not only on the input text but also the ground-truth label and other cached information. We present a unified view of robustness testing of summarization systems that is inspired by a diverse set of approaches to this problem \citep{grusky-etal-2018-newsroom,kedzie-etal-2018-content,jung-etal-2019-earlier,kryscinski-etal-2019-neural}.

\para{Models.}
We use model predictions for $7$ models from SummEval~\citep{fabbri2020summeval} on the CNN/DailyMail~\citep{hermann2015teaching} test dataset: (i) {\sc Lead-3}, which uses the $3$ leading sentences as a summary, (ii) {\sc Neusum}~\citep{Zhou2018NeuralDS}, (iii) {\sc Banditsum}~\citep{Dong2018BanditSumES}, (iv) {\sc Jecs}~\citep{Xu2019NeuralET}, (v) {\sc T5}~\citep{Raffel2020ExploringTL},  (vi) {\sc Bart}~\citep{Lewis2020BARTDS}, (vii) {\sc Pegasus}~\citep{Zhang2020PEGASUSPW}.

\para{Slices.}
Below, we define several heuristics for identifying subpopulations of summarization datasets for robustness testing. See Appendix~\ref{app:summarization} for additional details.

\begin{itemize}[leftmargin=*]
    \item {\it abstractiveness} is the degree to which the reference summary requires consolidating and reframing content from the source document~\citep{grusky-etal-2018-newsroom}. Summaries range from extractive, where a subset of sentences is directly selected from the source document to abstractive. 
    \item {\it distillation} is the degree to which the reference summary discards content from the source document. Highly distilled summaries require models to carefully select what to present in the summary. 
    \item {\it position} is the average location---in the source document---of where information in the summary comes from. High positions require models to use information that appears later in the source document.
    \item {\it dispersion} is the degree to which the reference summary uses content that is distributed broadly across the article versus concentrated in a particular region. High dispersion requires the method to attend to multiple parts of the source document to consolidate information.
    \item {\it ordering} is the similarity with which content in the source and reference summary are ordered. Summaries that change or reverse the ordering of content require models to reason over how to best present contextual information.
\end{itemize}
We also consider slices based on the length of the source document and the number of contained entities, which serve as proxy measures for the complexity of content to be summarized.

\begin{figure*}[!t]
    \centering
    \includegraphics[width=\textwidth]{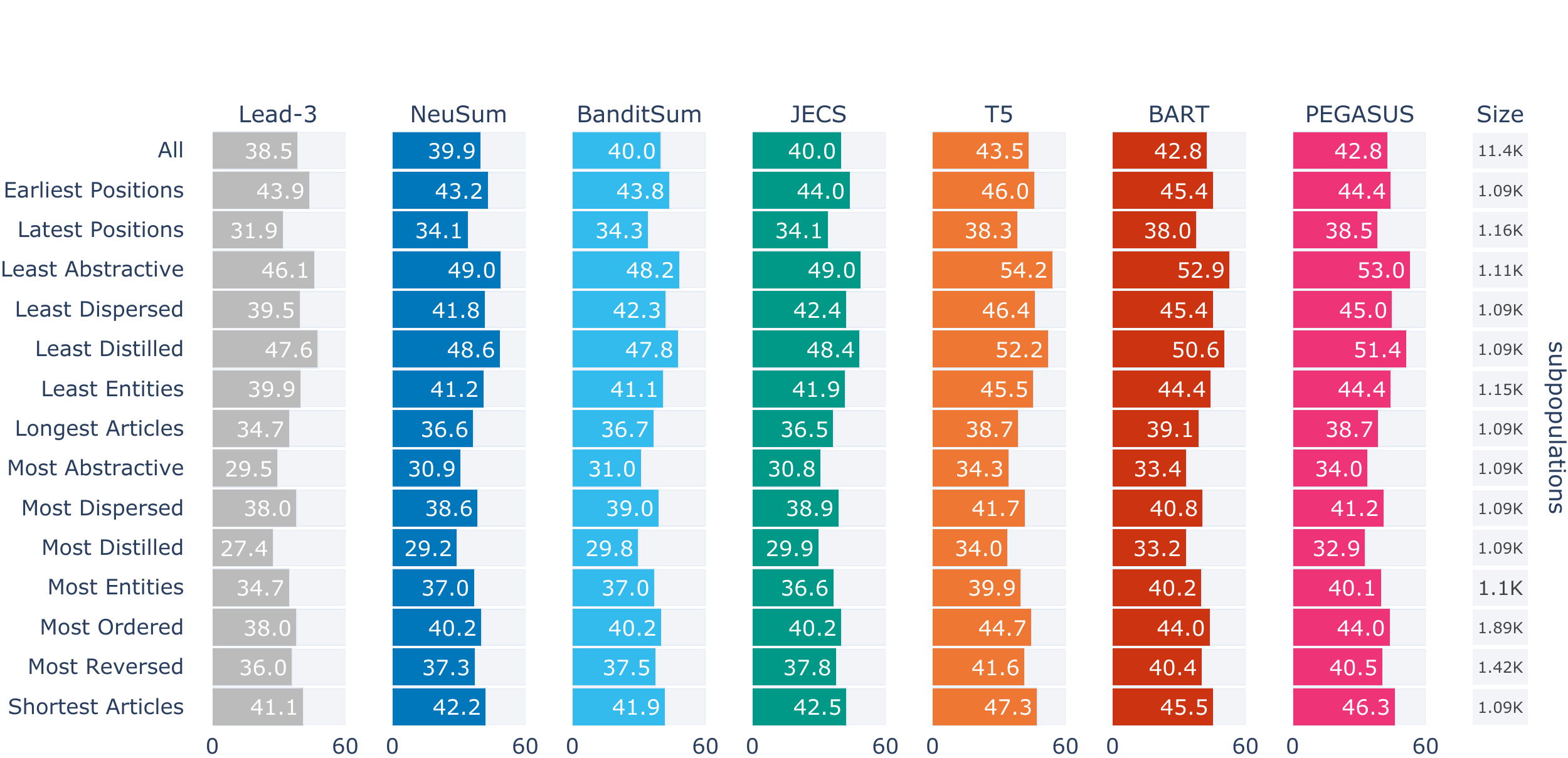}
    \caption{Robustness Report for summarization on CNN-DailyMail. Performance reported using the ROUGE-1 metric.}
    \label{fig:summarization-report}
\end{figure*}

\subsubsection{Analysis on CNN/DailyMail}
We include a Robustness Report in Figure~\ref{fig:summarization-report}, and describe results below.

\para{Models struggle to abstract and distill.} All models perform worst on the ``most distilled" subpopulation, i.e. on examples where the model must discard a large amount of information in the article to construct a summary. Models also struggle on the examples that required the most abstraction. In contrast, both extractive and abstractive models excel on extractive examples (``least abstractive"). 

\para{Abstractive models have less positional bias.} Extractive models have large gaps in performance between examples where the summary can be constructed using the early (``earliest positions") vs. late (``latest positions") of the article e.g. all extractive models have a $9+$ point gap between these subpopulations. Abstractive models have smaller gaps, e.g. {\sc Pegasus} has a gap of only $5.9$ points.

\para{Errors are highly correlated.} All summarization models, whether extractive or abstractive, degrade and improve on the same populations of data. 
This is surprising, since these models use quite different prediction mechanisms e.g. abstractive models like T5 appear to offer no relative advantage on the ``most abstractive" examples compared to the Lead-3 baseline (both models are $9$ points worse than their overall performance). 
We note that the development of reliable evaluation metrics in summarization continues to be an active area of research, and it is likely that current evaluation metrics are unable to capture some meaningful differences that may exist.

%% file: source/relatedwork.tex
\section{Related Tools and Work}
\label{sec:related}
Our work is related to many machine learning research areas including AutoML, Ethical ML, Interpretable ML, as well as Error Analysis.

\para{AutoML.} Automated machine learning is an area of research that focuses on creating tools that help remove the manual efforts in building machine learning systems~\citep{snoek2012practical}.  Traditionally, these have focused on data wrangling, feature and model selection, and hyperparameter optimization. More recently with hardware acceleration, AutoML has expanded to include neural architecture search (NAS)~\citep{pham2018efficient}. Although AutoML aims to provide tools for efficient and robust models, it only focuses on training and not evaluations~\citep{feurer2015efficient}. \RG{} on the other hand focuses on removing the manual effort in evaluations of machine learning models across a suite of robustness tests.

\para{Ethical ML.} There exist reporting and analysis tools developed for ethical, fair and responsible use of ML. The Model Cards toolkit~\citep{mitchell2019model} is an example of a reporting toolkit. Examples of analysis tools include the What-if Tool~\citep{wexler2019if}, FairVis~\citep{cabrera2019fairvis}, and FairSight~\citep{ahn2019fairsight}. Although these toolkits provide support to report and identify biases in ML models, it is not obvious how and which biases should be tested for. Their scope is also limited to ethics. \RG{} is a general-purpose toolkit that supports both reporting and analysis. It provides tools for evaluating robustness while mapping out the various dimensions that a user should consider for their use case.

\para{Interpretable ML.} Interpreting ML models enables a better understanding of their behavior. There exist several tools and frameworks for general purpose interpretability, including the recent Language Interpretability Tool (LIT)~\citep{tenney2020language}, IBM's AI Explainability 360~\citep{aix360-sept-2019}, AllenNLP Interpret~\citep{wallace2019allennlp}, InterpretML~\citep{nori2019interpretml}, Manifold~\citep{zhang2018manifold}, and Pytorch Captum~\citep{captum}. DiCE is a tool focused on explanations using counterfactuals~\citep{mothilal2020dice}. 
Interpretability and robustness are both desirable but different properties for ML models. Interpretability tools not only have different objectives but also different sets of features that are complementary to \RG{} (e.g., interpreting or providing causal explanations for a particular prediction after \RGabbrv~identifies a generalization problem in the model). Many of these tools focus on interactive visualization, which limits their scope to interpreting small numbers of examples and makes their results difficult to reproduce. This also makes their use susceptible to subjectivity and selection bias. By contrast, \RG{} can scale to large datasets (e.g., $100,000$ Wikipedia examples in Section~\ref{sec:nel}) with ease. Testbenches provide reproducibility to the analyses conducted in \RGabbrv. %

\para{Error Analysis.} Tools for error analysis help users in understanding where their models are failing. Errudite~\citep{Wu2019ErruditeSR} supports users in exploring subpopulations of their data, while CrossCheck~\citep{Arendt2020CrossCheckRR} and Manifold~\citep{zhang2018manifold} focus on visualization and analysis for model comparison. \RG{} is complementary to these tools in that it enables users to understand likely performance degradations and preempt those before they become errors.

%% file: source/conclusion.tex
\section{Conclusion}
We introduced \RG, an evaluation toolkit that supports a broad set of evaluation idioms, and can be used for collaboratively building and sharing evaluations and results. To address challenges faced by practitioners today, we embedded \RG{} into the Contemplate $\rightarrow$ Create $\rightarrow$ Consolidate continual evaluation loop. Our results suggest that \RG{} is a promising tool for researchers and practitioners.

%% file: source/appendix.tex
\appendix
\section{Appendix}
\subsection{Commercial System Case Study}
\label{app:study}
\paragraph{Pre-study questionnaire.}We asked the user to fill out the first questionnaire before the user study session and the second one after the session. The pre-study form included the following questions: which NLP task is the team working on (sentiment, dialog, question answering, natural language inference, machine translation, language modeling, summarization, and others), what metrics does the team use for evaluating their models (accuracy, P/R/F1, exact match, BLEU, ROUGE, or other generation metrics, and others), how they evaluate robustness (standard val/test datasets, out-of-distribution examples or datasets for generalization testing, axiomatic bias tests, adversarial attacks, and model cards). The form also asked the user to rate on a Likert scale of 1-5, 1 being strongly disagree and 5 being strongly agree, the following two statements: ``I would like to evaluate the robustness of my model more thoroughly than I do today.'' and ``I would benefit from having a library that gives me the tools to evaluate the robustness of my models.''
They rated the aforementioned agreement statements as 4/5 and 5/5 respectively.

\paragraph{Post-study questionnaire.}The post-study questionnaire evaluated \RG{} in terms of ease of use and how likely they are to incorporate the gym in their workflow on a Likert scale of 1-5, 1 being ``very unlikely'' and 5 being ``very likely''. At the end of the study, the team rated ``very likely'' for both ease of use and eagerness for using \RG{} as part of their workflow. 

One question was about rating the usefulness of the 4 evaluation idioms in the study. The team rated subpopulations and adversarial attacks as 5/5, transformations as 4/5, and eval sets as 3/5 on a scale of 1-5, 1 being ``not useful'' and 5 being ``very useful''. For the key takeaways of the study, the team found subpopulation slices as being ``very insightful''. They were very happy that they could first use adversarial attacks to probe for vulnerabilities and then use augmentations to fix them all in one tool.

\subsection{Named Entity Linking}

{\bf AIDA.} For the AIDA test-b dataset, we follow~\citep{Fvry2020EmpiricalEO} to split each passage in the dataset into examples. Each example corresponds to one sentence in the passage, pre-pended with the leading sentence that the passage starts with as context. We ignore predictions over the context sentence when calculating metrics.

\subsection{Summarization}
\label{app:summarization}
We describe the summarization slices in more detail below.

\para{Abstractiveness.}
The degree to which the reference summary is abstractive versus extractive \citep{grusky-etal-2018-newsroom}, based on the proportion of n-grams in the reference summary that are \textit{not} in the article. 
Formally, we define the abstractiveness of a summary $S$ given an article $A$ as: 
\begin{equation*}
\texttt{abstractiveness}(A,S) = \\ 1 - \texttt{rouge}_{\texttt{precision}}(A,S)
\end{equation*}

\noindent based on several variations of Rouge (Rouge-1, Rouge-2, Rouge-L). Note that $\texttt{rouge}_{\texttt{precision}}(A,S)$ equals the proportion of n-grams in the reference summary that are also in the article. The abstractiveness metric is essentially the complement of the Extractive Fragment Coverage metric introduced in \citet{grusky-etal-2018-newsroom}.

\para{Distillation.} The degree to which the reference summary is distilled from a larger quantity of content, based on the proportion of n-grams in the article that do \textit{not} appear in the reference summary:
\begin{equation*}
\texttt{distillation}(A,S) = 1 - \texttt{rouge}_{\texttt{recall}}(A,S)
\end{equation*}
Note that $\texttt{rouge}_{\texttt{recall}}(A,S)$ equals the proportion of n-grams in the article that appear in the reference summary. 

\para{}We also consider $3$ fine-grained metrics that rely on the similarities between sentences in the article and sentences in the reference summary. For these metrics, we define a sentence-similarity matrix $M$, where $M_{i,j}$ is a similarity score (e.g. Rouge-1) between sentence $a_i$ in the article and sentence $s_j$ in the summary. We provide the sentence-similarity matrix $M$ as a built-in abstraction in \RG{}, from which a variety of metrics may be decoded. Sharing this abstraction not only reduces code reuse, but also lowers the computational cost when performing multiple evaluations.

\para{}We also define a \texttt{match} function, which returns the index $i$ of the sentence in the article with greatest similarity to the summary sentence $s_j$:
\begin{equation*}
\texttt{match}(j) = \argmax_i(M_{i,j})
\end{equation*}

\noindent
Based on these formalisms, we define $3$ metrics:

\para{Position.} The mean position of the matched sentences in the article:
\begin{equation*}
\texttt{position}(A,S) = \sum_{j=1}^N\texttt{match}(j) \big / N
\end{equation*}

\noindent
This metric is inspired by previous work showing that summarization models may be biased toward sentences at the beginning of an article \citep{kedzie-etal-2018-content,jung-etal-2019-earlier,kryscinski-etal-2019-neural}.

\para{Dispersion.} The degree to which summary sentences match content that is distributed broadly across the article versus concentrated in a particular region. 
We define dispersion as the variance of the position of the matched sentences:
\begin{equation*}
\texttt{dispersion}(A,S) = \sum_{j=1}^N(\texttt{match}(j) - \mu)^2  \big / N
\end{equation*}

\noindent where $\mu$ is the mean match position, which equals $\texttt{position}(A,S)$, defined earlier. This metric is related to Extractive Fragment Density \citep{grusky-etal-2018-newsroom}, which measures the degree to which extracted text in the summary comes from a contiguous sequence versus being broadly sourced from the article. 

\para{Order.} The similarity in ordering between the summary sentences and the matched article sentences. Specifically, we compute the Spearman rank correlation between the positions of sentences in the reference summary and the positions of their matched counterparts in the article:
\begin{equation*}
\texttt{order}(A,S) = \texttt{spearman}(
(\texttt{match}(j))_{j=1}^N, (j)_{j=1}^N) 
\end{equation*}

This metric is inspired by prior work in summarization evaluation that studied the effects of shuffling sentences in the source article, revealing a significant degradation in performance in news articles compared to other domains \citep{kedzie-etal-2018-content}.

\subsection{Code and Reports}
\para{Code.} We provide example code snippets for \RG{} in Tables \ref{tab:code-cachedops} (\cachedop), \ref{tab:code-slicebuilders} (\slicebuilder), and \ref{tab:code-reporting} (\testbench, \report), below.

\begin{figure*}[h]
\begin{center}
\includegraphics[width=\linewidth]{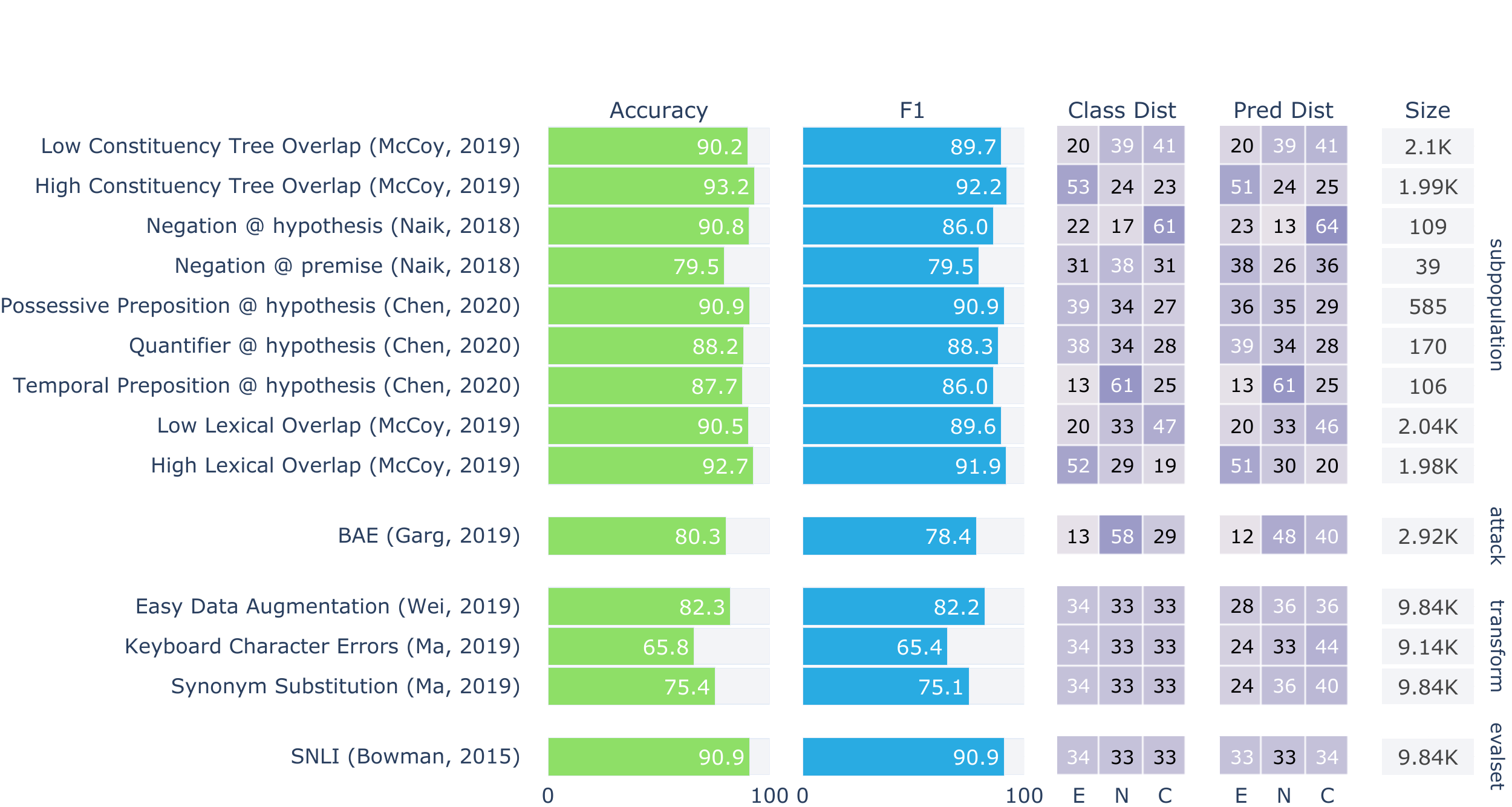}
\end{center}
\caption{Robustness report for textattack/bert-base-uncased-snli model on SNLI dataset. The report lays out scores for each evaluation, broken out by category. Citations: \citep{chen2019slice,
naik2018stress,
McCoy2019RightFT,
wei2019eda,
nlpaug,
bowman2015large}.
\\
\\
\textit{Note: the \LaTeX\ figure and caption above is auto-generated using ``report.latex()".}
}
\label{fig:latex-report}
\end{figure*}

\para{\LaTeX\ Report.} Figure~\ref{fig:latex-report} is an example of a report generated in a \LaTeX\ format. The code for the figure was auto-generated and the figure was simply included in the appendix.

\begin{table*}[!h]
    \lstset{linewidth=8.2cm}
    \centering
    \scriptsize
    \begin{tabular}{p{0.02\linewidth}p{0.08\linewidth}p{0.2\linewidth}p{0.7\linewidth}}
    \toprule
        \textbf{Goal} & &&\textbf{Code Snippet} \\
        \midrule
        \multirow{45}{*}{\rotatebox[origin=c]{90}{Caching}}&
        \multirow{20}{*}{Create}&
        Create Spacy cached operation
        &   
        \begin{lstlisting}
        spacy = Spacy()
        \end{lstlisting}
        \\&&
        Create Stanza cached operation
        &   
        \begin{lstlisting}
        stanza = Stanza()
        \end{lstlisting}\\
        &&
        Create a custom cached operation
        &   
        \begin{lstlisting}
        cachedop = CachedOperation(
            apply_fn=my_custom_fn,  
            identifier=Identifier('MyCustomOp'),
        )
        \end{lstlisting}
        \\
        &&
        Run a cached operation
        &   
        \begin{lstlisting}
        dataset = cachedop(dataset, columns)
        \end{lstlisting}
        \\
        
        \cmidrule{3-4}
        
        &\multirow{20}{*}{Retrieve}&
        Retrieve all Spacy info cached
        &   
        \begin{lstlisting}
         Spacy.retrieve(dataset, columns)
        \end{lstlisting}\\
        
        &&
        Retrieve Spacy tokens     
        &   
        \begin{lstlisting}
         Spacy.retrieve(batch, columns, 'tokens')
        \end{lstlisting}\\
        
        &&
        Retrieve Stanza entities     
        &   
        \begin{lstlisting}
         Stanza.retrieve(
            batch, 
            columns, 
            Stanza.entities
        )
        \end{lstlisting}\\
        
        &&
        Retrieve any cached operation info after processing
        &   
        \begin{lstlisting}
         CachedOperation.retrieve(
            batch, 
            columns, 
            my_proc_fn, 
            'MyCustomOp'
        )
        \end{lstlisting}\\        
        \bottomrule
    \end{tabular}
    \caption{Code for the \cachedop\ abstraction in \RG.}
    \label{tab:code-cachedops}
\end{table*}

\begin{table*}[]
    \lstset{linewidth=8.2cm}
    \centering
    \scriptsize
    \begin{tabular}{p{0.02\linewidth}p{0.12\linewidth}p{0.25\linewidth}p{0.61\linewidth}}
    \toprule
        \textbf{Goal} & &&\textbf{Code Snippet} \\
        \midrule
        \multirow{58}{*}{\rotatebox[origin=c]{90}{Slice Building}}&
        \multirow{20}{*}{Subpopulations} 
        & Create a subpopulation that generates three slices based on raw lengths in $[0, 10]$, $[10, 20]$ and $[20, \infty)$  & 
        \begin{lstlisting}
        length_sp = Length(
            [(0, 10), (10, 20), (20, np.inf)]
        )
        \end{lstlisting}\\
        
        && Create a subpopulation that generates two slices based on bottom $10\%$ and top $10\%$ length percentiles  & 
        \begin{lstlisting}
        length_sp = Length(
            [('0%', '10%'), ('90%', '100%')]
        )
        \end{lstlisting}\\

        && Create a custom subpopulation by binning the outputs of a scoring function 
        &
        \begin{lstlisting}
        custom_sp = ScoreSubpopulation(
            [('0%', '10%'), ('90%', '100%')], 
            my_scoring_fn
        )
        \end{lstlisting}\\
        
        \cmidrule{3-4}
        
        &\multirow{15}{*}{Transformations}&Create EasyDataAugmentation&
        \begin{lstlisting}
        eda = EasyDataAugmentation()
        \end{lstlisting}\\
        
        &&Create any NlpAug transformation &
        \begin{lstlisting}
        nlpaug_trans = NlpAugTransformation(
            pipeline=nlpaug_pipeline
        )
        \end{lstlisting}\\
        &&Create a custom transformation &
        \begin{lstlisting}
        custom_trans = Transformation(
            Identifier('MyTransformation'), 
            my_transformation_fn
        )
        \end{lstlisting}\\
        
        \cmidrule{3-4}
        
        &\multirow{5}{*}{Attacks}&Create TextAttack recipe &
        \begin{lstlisting}
        attack = TextAttack.from_recipe(recipe, model)
        \end{lstlisting}\\
        
        \cmidrule{3-4}
        
        &\multirow{5}{*}{Evaluation Sets}
        & Create a slice from a dataset & 
        \begin{lstlisting}
        sl = Slice(dataset)
        \end{lstlisting}\\
        
        \cmidrule{3-4}
        
        &\multirow{7}{*}{Slice Builders}&Run any SliceBuilder & 
        
        \begin{lstlisting}
        dataset, slices, membership = slicebuilder(
            batch_or_dataset=dataset, 
            columns=columns,
        )
        \end{lstlisting}\\
        
        \bottomrule
    \end{tabular}
    \caption{Code for the \slicebuilder\ abstraction in \RG.}
    \label{tab:code-slicebuilders}
\end{table*}

\begin{table*}[]
    \lstset{linewidth=8.2cm}
    \centering
    \scriptsize
    \begin{tabular}{p{0.03\linewidth}p{0.1\linewidth}p{0.2\linewidth}p{0.67\linewidth}}
    \toprule
        \textbf{Goal} & &&\textbf{Code Snippet} \\
        \midrule
        \multirow{50}{*}{\rotatebox[origin=c]{90}{Reporting}}&\multirow{25}{*}{Testbench}&Create a testbench & 
        \begin{lstlisting}
        testbench = TestBench(
            identifier=Identifier('MyTestBench'), 
            version='0.1.0'
        )
        \end{lstlisting}\\
        
        && Add slices to testbench & 
        \begin{lstlisting}
        testbench.add_slices(slices)
        \end{lstlisting}\\
        
        && Fuzzy search testbench for slices & 
        \begin{lstlisting}
        top_k_matched_slices = testbench.search('len')
        \end{lstlisting}\\
        
        && Bump testbench minor version & 
        \begin{lstlisting}
        testbench.bump_minor()
        \end{lstlisting}\\
        
        && Save and load a testbench & 
        \begin{lstlisting}
        testbench.save(path)
        testbench.load(path)
        \end{lstlisting}\\
        
        \cmidrule{3-4}
        
        &\multirow{20}{*}{Report}&&\\
        && Evaluate model on slices and generate report & 
        \begin{lstlisting}
        testbench.create_report(model)
        \end{lstlisting}\\
        
        && Create a custom report & 
        \begin{lstlisting}
        report = Report(
            dataframe_with_metrics, 
            report_columns,
        )
        \end{lstlisting}\\
        
        && Generate figure from report & 
        \begin{lstlisting}
        figure = report.figure()
        \end{lstlisting}\\
        
        && Generate \LaTeX report & 
        \begin{lstlisting}
        latex = report.latex()
        \end{lstlisting}\\
        \bottomrule
    \end{tabular}
    \caption{Code for the \testbench\ and \report\ abstractions in \RG.}
    \label{tab:code-reporting}
\end{table*}

